\documentclass[journal]{IEEEtran}
\usepackage{amsmath}
\usepackage{amssymb}
\usepackage{booktabs}
\usepackage{graphicx}
\usepackage{subfigure}
\usepackage{float}
\usepackage{url}
\usepackage{multirow}
\usepackage{amsthm}
\usepackage[linesnumbered,boxed,ruled,commentsnumbered,noend]{algorithm2e}

\newtheorem{theorem}{Theorem}

\theoremstyle{definition}
\newtheorem{remark}{Remark}

\newtheorem{definition}{Definition}

\ifCLASSINFOpdf

\else

\fi

\usepackage{xcolor}

\begin{document}
\title{Multi-UAV Deployment in Obstacle-Cluttered Environments with LOS Connectivity}

\author{Yuda Chen and Meng Guo
\thanks{
	The authors are with the State Key Laboratory for Turbulence and
	Complex Systems, Department of Mechanics and Engineering Science,
	College of Engineering, Peking University, Beijing 100871, China.
        This work was supported by the National Natural Science Foundation
    of China (NSFC) under grants 62203017, T2121002, U2241214;
    and by the Fundamental Research Funds for the central universities.
    Contact: {\tt\small meng.guo@pku.edu.cn}.
}
}

% \author{\cdots.lastname \thanks{\cdots} \thanks{\cdots} }

% \markboth{Journal of \LaTeX\ Class Files,~Vol.~14, No.~8, August~2015}%
% {Shell \MakeLowercase{\textit{et al.}}: Bare Demo of IEEEtran.cls for IEEE Journals}

\maketitle

\pagestyle{empty}  % no page number for the second and the later pages
\thispagestyle{empty} % no page number for the first page

\begin{abstract}
	A reliable communication network is essential for multiple UAVs operating within obstacle-cluttered environments,
	where limited communication due to obstructions often occurs.
	A common solution is to deploy intermediate UAVs to relay information via a multi-hop network,
	which introduces two challenges:
	(i) how to design the structure of multi-hop networks;
	and (ii) how to maintain connectivity during collaborative motion.
	To this end, this work first proposes an efficient constrained search method based on the minimum-edge RRT$^\star$ algorithm, to find a spanning-tree topology that requires a less number of UAVs for the deployment task.
	To achieve this deployment,
	a distributed model predictive control strategy is proposed for the online motion coordination.
	It explicitly incorporates not only the inter-UAV and UAV-obstacle distance constraints,
	but also the line-of-sight (LOS) connectivity constraint.
	These constraints are well-known to be nonlinear and often tackled by various approximations.
	In contrast, this work provides a theoretical guarantee
	that
	%the desired topology can be reached during which
	all agent trajectories are ensured to be collision-free with a team-wise LOS connectivity at all time.
	Numerous simulations are performed in 3D valley-like environments,
	while hardware experiments validate its dynamic adaptation
	when the deployment position changes online.
\end{abstract}

% \fi
\IEEEpeerreviewmaketitle

%%%%%%%
%%%%%%%

\section{Introduction}

Fleets of unmanned aerial vehicles (UAVs) have been widely deployed to accomplish
collaborative missions, e.g., for exploration, inspection, search and rescue~\cite{Chung2018}.
In many applications,
the UAVs need to keep a reliable communication with a ground station to transfer data streams and monitor operation status of the fleet.
This however can be challenging due to the limited communication range and obstructions from obstacles.
An effective solution as also adopted
in~\cite{Nouyan2009,Dorigo2013,Stephan2017}
is to deploy more UAVs as relay nodes to form a multi-hop
communication network,
such that all agents can transmit and receive data
reliably with the ground station.
Two key challenges arise with this approach, i.e.,
(i) how the embedded multi-hop network should be designed,
including the topology and the embedded positions of each node;
(ii) how to control the UAV fleet to form this network collaboratively,
while maintaining the team-wise connectivity and avoiding collisions
within an obstacle-cluttered environment.

\subsection{Related Work}
Regarding the design of communication network,
linear graphs are proposed in~\cite{Schouwenaars2006,Derbas2014,
	Varadharajan2020}
for exploration and surveillance tasks in complex environments.
Such structure is simple to design but rather limited, as it only allows the tailing agent to actively perform a task while the other agents serve as relay nodes.
Another common topology is called spanning trees as adopted in~\cite{Majcherczyk2018,Luo2020,Yi2021},
where all agents as leaves of this tree can execute multiple tasks simultaneously.
However, these aforementioned works often assume a redundant number of agents
without estimating or minimizing the number of agents that is required to accomplish the given set of tasks.
Furthermore, the communication model considered in these work is often simplified to be range-based,
i.e., any two agents within a certain distance can exchange information freely.
However, as discussed in~\cite{Goldsmith2005}, obstructions due to obstacles can severely degrade the communication quality and the transmission rate.
Thus, as also motivated in~\cite{Esposito2006,Anisi2010,Boldrer2021},
the line-of-sight (LOS) model is a more practical choice for approximating the inter-agent communication constraints in obstacle-cluttered environments.
More concretely, the work in~\cite{Boldrer2021} proposes the notion of visibility-weighted minimum spanning tree (MST), which leads to a more robust path compared to~\cite{Luo2020} under LOS constraints.
However, it does not appropriate for the motion coordination among the agents with complex dynamics like UAVs.

\begin{figure}[t!]
	\centering
	\includegraphics[height=0.34\linewidth]{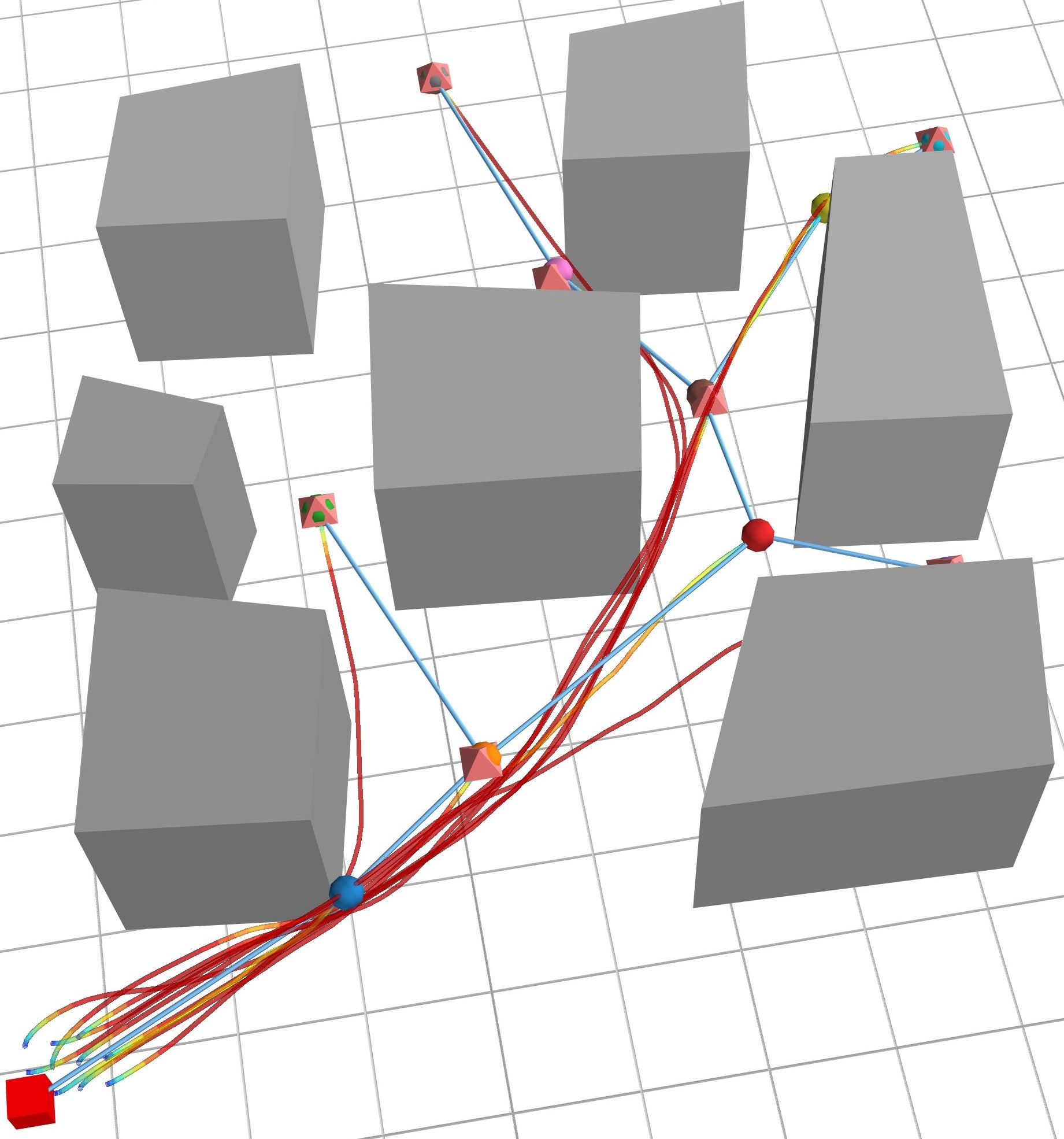}
	\hspace{0.1in}
	\includegraphics[height=0.34\linewidth]{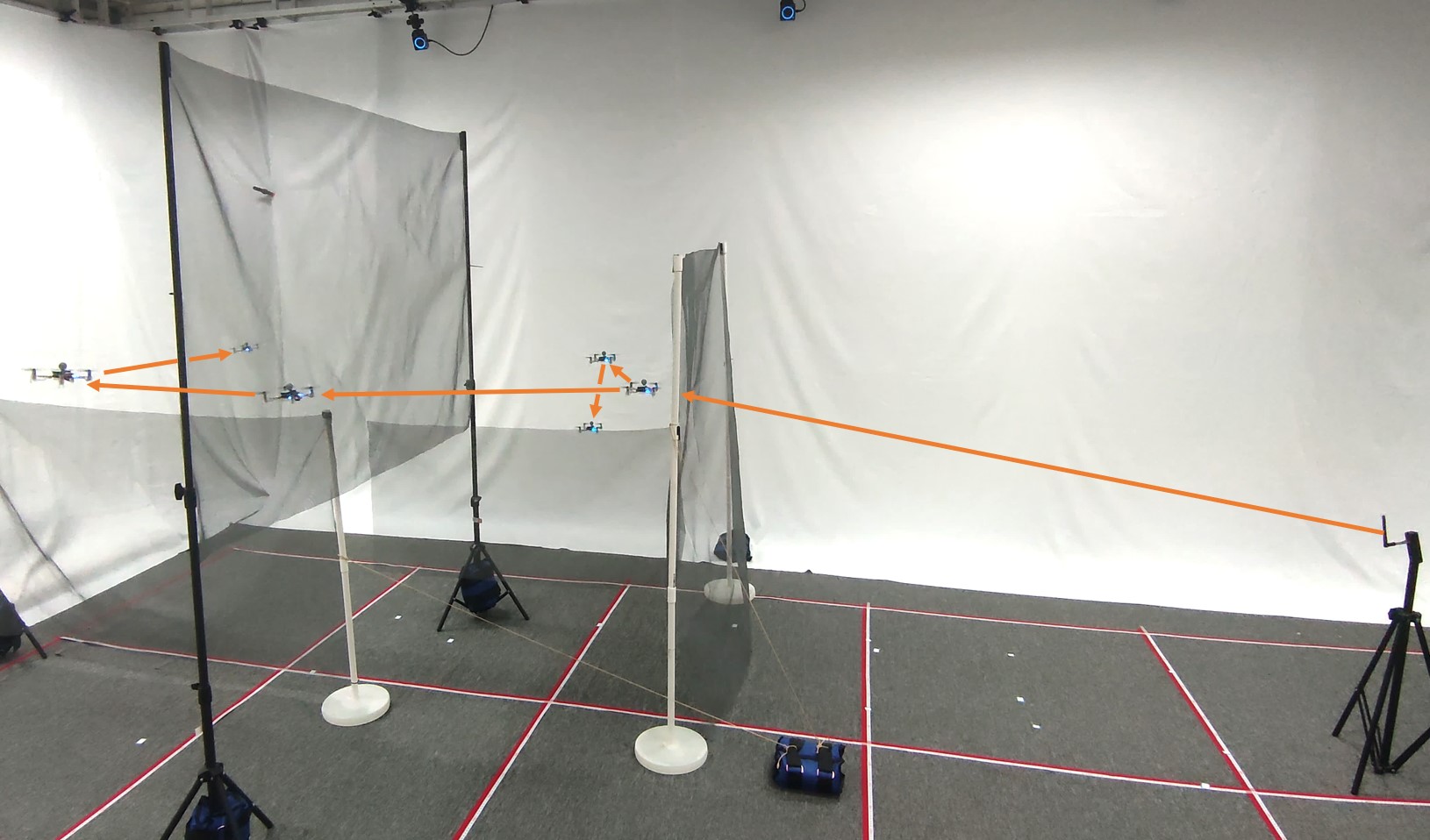}
	\caption{
		\textbf{Left}:
		Ten agents (colored spheres) are deployed from the ground station (red cubic) to seven targets (pink diamonds).
		The optimized network topology is shown in blue lines,
		which is kept during the motion at all time.
		\textbf{Right}: Experiment of $6$ quadrotors cooperatively searching two targets.}
	\label{fig:overall}
\end{figure}

On the other hand,
it is a non-trivial problem to maintain the designed communication network during collaborative motion,
especially when avoiding collisions between agents and obstacles.
Numerous methods have been proposed to address these constraints simultaneously.
For instances, artificial forces are designed in~\cite{Feng2015,Paolo2013},
e.g., repulsive forces for obstacles and other agents;
and attractive forces for the targets and connected agents.
Lyapunov-based nonlinear controllers are proposed in~\cite{zavlanos2008distributed,zavlanos2011graph}
mostly for obstacle-free environments.
Control barrier certificates are proposed in~\cite{Luo2020,Wang2016}
for both  constraints of connectivity maintenance and collision avoidance.
Despite of their intuitiveness, 
these methods often lack theoretical guarantee on the safety when encountering control and state constraints.
Furthermore, the LOS restriction is particularly difficult to incorporate due to its non-convexity.
A geometric approach is proposed in~\cite{Boldrer2021} based on the Loyd algorithm,
which drives the agents towards the intersection of neighboring communication zones.
It however neglects the dynamic constraints such as limited velocity and acceleration.
Optimization-based methods such as~\cite{Schouwenaars2006,Caregnato2022} directly formulate these constraints as mixed integer programs (MIP).
However, the complexity increases drastically with the number of agents and obstacles,
yielding it impractical for real-time applications.

\subsection{Our Approach}

To tackle these limitations, this work proposes an integrated framework of network optimization
and motion coordination for deploying UAV fleets in obstacle-cluttered environments.
It generates an embedded network topology as spanning trees via the minimum-edge RRT$^\star$ algorithm,
subject to the LOS restrictions.
During online execution, a collaborative motion coordination strategy is proposed based on the
distributed model predictive control (MPC) that maintains the designed network topology
and generates collision-free trajectories.
In particular, the LOS constraints as well as the requirement of collision avoidance are explicitly formulated as linear constraints by restricting the UAVs in constructed convex polyhedras.
It has been shown that the distributed MPCs can be solved online efficiently,
with a theoretical guarantee on its feasibility.
Effectiveness of the proposed framework is validated in numerous simulations and hardware experiments.

The main contributions are two-fold:
(i)  Different from the common self-organizing topology
in~\cite{Varadharajan2020,Luo2020,Boldrer2021},
the proposes algorithm of topology optimization optimizes the number of agents required for the task;
(ii) The proposed motion control strategy based on
distributed MPCs has a theoretical guarantee on the network connectivity and collision-free trajectories.
In comparison with MILP-based methods in \cite{Caregnato2022} which handles LOS restrictions, 
our method reduces the planning time by $90\%$ and exhibits online adaptation
towards dynamic target positions. 
%To the best of our knowledge, 
%it is the first work that addresses the LOS connectivity constraint within the framework of distributed MPC.

\section{Problem Formulation}
\subsection{Robot Model}
As illustrated in Fig.~\ref{fig:overall},
there is a fleet of $N>0$ autonomous UAVs within the 3D workspace~$\mathcal{W}\subset \mathbb{R}^{3}$.
The dynamic model of each agent follows the standard double-integrator model as
$\dot{x}^i(t) = [v^i(t), u^i(t)]$,
where the state $x^i(t)=[p^i(t),\, v^i(t)]$;
$p^i(t), v^i(t), u^i(t) \in \mathbb{R}^3$ represent the position, velocity and control input of agent~$i$ at time~$t\geq 0$,
respectively, $\forall i\in \mathcal{N} = \{1,\cdots,N\}$.
Moreover, the velocity and inputs of each agent~$i\in \mathcal{N}$
are restricted by
$\| \Theta_a u^i(t) \|_2 \le a_{\text{max}}, \; \| \Theta_v v^i(t) \|_2 \le v_{\text{max}}$, $\forall t$
where
$\Theta_v,\, \Theta_a$ are positive-definite matrices, and $v_{\text{max}},\, a_{\text{max}}>0$ are the maximum velocity and acceleration, respectively.
Note that the total number of agents $N$ is \emph{not} fixed,
rather optimized as part of this problem.

\subsection{Communication Constraints}
As the means to exchange information,
any pair of agents $i,j\in \mathcal{N}$ can direct communicate
as \emph{neighbors} if the following two conditions hold:
(i) their relative distance is bounded by the communication range~$d_c>0$,
i.e., $\|p^i(t)-p^j(t)\|_2 \leq d_c$;
(ii) the line-of-sight (LOS) between agents~$i$ and $j$
cannot be obstructed by obstacles, i.e.,
$\textbf{Line}(p^i(t),\, p^j(t)) \cap \mathcal{O} = \emptyset,$
where $\textbf{Line}(p^i(t),\, p^j(t))$ is the line segment
from position~$p^i(t)$ to position~$p^j(t)$;
and
$\mathcal{O}\subset \mathcal{W}$ is the volume occupied
by a set of convex-shaped static obstacles,
each of which is defined as the convex hull of a set of known vertices.

Consequently, an un-directed and time-varying
communication network $G(t)=(\mathcal{N},\, E(t))$
can be constructed given the neighboring relation above,
where $E(t)\subset \mathcal{N}\times \mathcal{N}$ for time~$t>0$.
In other words, $(i,\,j)\in E(t)$ if the position of agents~$i$ and $j$
at time~$t$ satisfies the two conditions above.
Network $G(t)$ is called \emph{connected} at time~$t$
if there exists a path between any two nodes in $G(t)$.

\subsection{Collision Avoidance}

For safety, each agent~$i\in \mathcal{N}$
occupies a ball of radius $r_a>0$, i.e.,
$\mathcal{B}^i(t)=\{p\in \mathcal{W}\,|\, \|p-p^i(t)\|_2 \leq r_a\}$.
Thus, any pair of two agents~$i,j\in \mathcal{N}$
might collide with each other
if their relative distance is less than $2r_a$,
i.e., a collision happens when $ \| p^i - p^j \|_2 < 2 r_a$.
Furthermore, the agent~$i\in\mathcal{N}$ might collide with the static obstacles at time~$t>0$,
if their occupied volumes intersect,
i.e., $\mathcal{B}^i(t) \cap \mathcal{O} \neq \emptyset$ should be satisfied.

\subsection{Problem Formulation}

Lastly, there are $M>0$ target locations scattered within the free-space,
denoted by
$\mathcal{P}_\text{target} =  \left\{p_{\text{target},m},\,\forall m \in \mathcal{M} \right\}\subset \mathcal{W}\backslash \mathcal{O}$,
where $\mathcal{M}=\{1,\cdots,M\}$;
and each location contains potential tasks to be performed by the agents.
Additionally, assume that agent~$i\in \mathcal{N}$ starts from an initial position~$p^i_0$ in the freespace,
such that the fleet is collision-free initially
and the communication network~$G(0)$ is connected.
Thus, the considered problem is stated formally as a constrained optimization problem in the following:
\begin{equation}\label{eq:problem}
	\begin{split}
		&\textbf{min}_{\{u^i(t),\, T\}}\; N\\
		\textbf{s.t.}\quad & p^{i_m}(T)=p_{\text{target},m},\;\forall m;\\
		& \dot{x}^i(t) = [v^i(t),u^i(t)], \; \forall i,\;\forall t;\\
		& \| \Theta_a u^i(t) \|_2 \le a_{\text{max}}, \; \| \Theta_v v^i(t) \|_2 \le v_{\text{max}}, \; \forall i,\;\forall t;\\
		& G(t)\text{ is connected}, \;\forall t;\\
		& \mathcal{B}^i(t)\cap \mathcal{B}^j(t) = \emptyset, \; \mathcal{B}^i(t)\cap \mathcal{O} = \emptyset, \;\forall i,\;\forall j,\;\forall t;
	\end{split}
\end{equation}
where $\forall i$ stands for $\forall i\in\mathcal{N}$ (the same for $j$);
$\forall t$ stands for $\forall t\in [0, T]$;
$\forall m$ stands for $\forall m \in \mathcal{M}$;
$T>0$ is the termination time when every target position is reached;
$i_m\in \mathcal{N}$ is the agent that reaches target~$m\in \mathcal{M}$;
and the conditions above are the dynamic constraints, the communication connectivity,
and the safety constraints as described previously.
Note that the objective is to minimize the total number of agents required for this mission.

\section{Proposed Solution}

The proposed solution consists of two layers.
First, the communication network is optimized given the
workspace layout and the target locations, in order to minimize the team size.
Second, a collaborative control strategy is designed to drive the agents
towards the target positions while maintaining the desired topology,
subject to the collision avoidance and connectivity maintenance.

\subsection{Optimization of the Communication Network} \label{sub:network-formulation}

\begin{figure}[t!]
	\centering
	\includegraphics[width=0.85\linewidth]{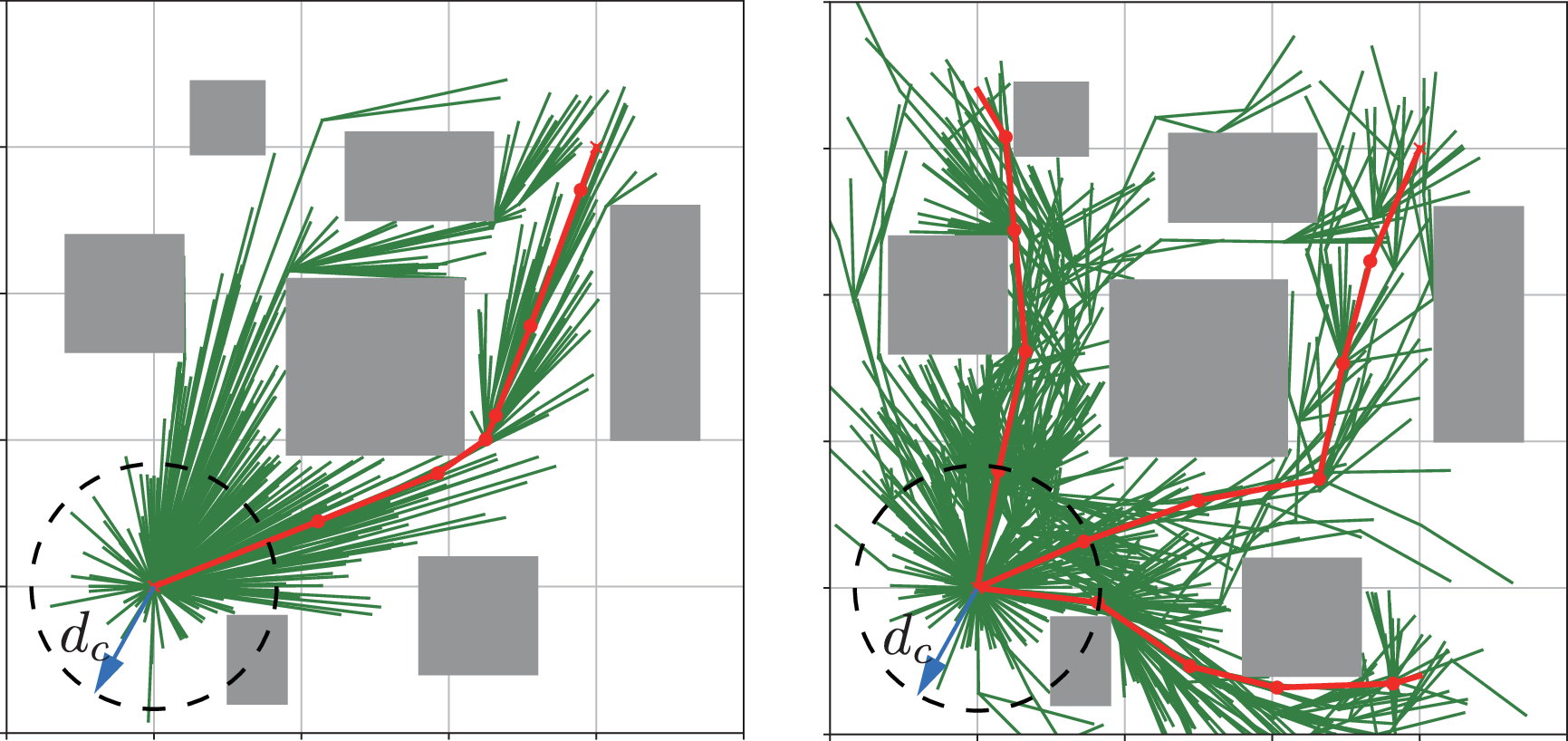}
	\caption{ \textbf{Left}: the RRT$^\star$ in~\cite{Gammell2014}
		finds the shortest path between any two points.
		\textbf{Right}: minimum-edge RRT$^\star$ is used to find paths from
		a staring point to multiple target points, each with a minimum-edge path.}
	\label{fig:discrete-RRT}
\end{figure}

As a building block of the proposed optimization algorithm for the communication network,
the RRT$^\star$ proposed as Algorithm~1 of~\cite{Gammell2014}
is modified to find a list of paths from a starting point to a set of target points
with a minimum number of edges along each path.
More specifically,
the following three modifications are made:
(i) the paths towards more than one target points can be concurrently found;
(ii) the cost of each edge is augmented by a large penalty to minimize the
number of edges along a path;
(iii) new samples are generated within the distance~$d_c$ of
an existing sample to reflect the limited communication range.
As a result,
a path with minimum edges can be found between the starting point and
any target point.
The above procedure is illustrated in Fig.~\ref{fig:discrete-RRT}.
However, these paths could be further improved by increasing the number
of \emph{shared} edges among them.
To begin with, a spanning-tree topology is defined as follows.

\begin{definition}
	An \emph{embedded} spanning tree is defined as a set of 3-tuples:
	$\mathcal{T}=\{V_0, V_1, \cdots \}$,
	where each node~$V_i = (i,\, p_i,\, I_i)$,
	with index~$i\in \mathbb{Z}$, position~$p_i\in \mathcal{W}$
	and parent index~$I_i\in \mathbb{Z}$ of node~$i$, $\forall i=0,1,\cdots$.
	Note that~$V_0=(0,\, p_{g},\,\emptyset)$ is the root node associated with the ground station. \hfill $\blacksquare$
\end{definition}

\begin{figure}[t]
	\centering
	\includegraphics[width=0.95\linewidth]{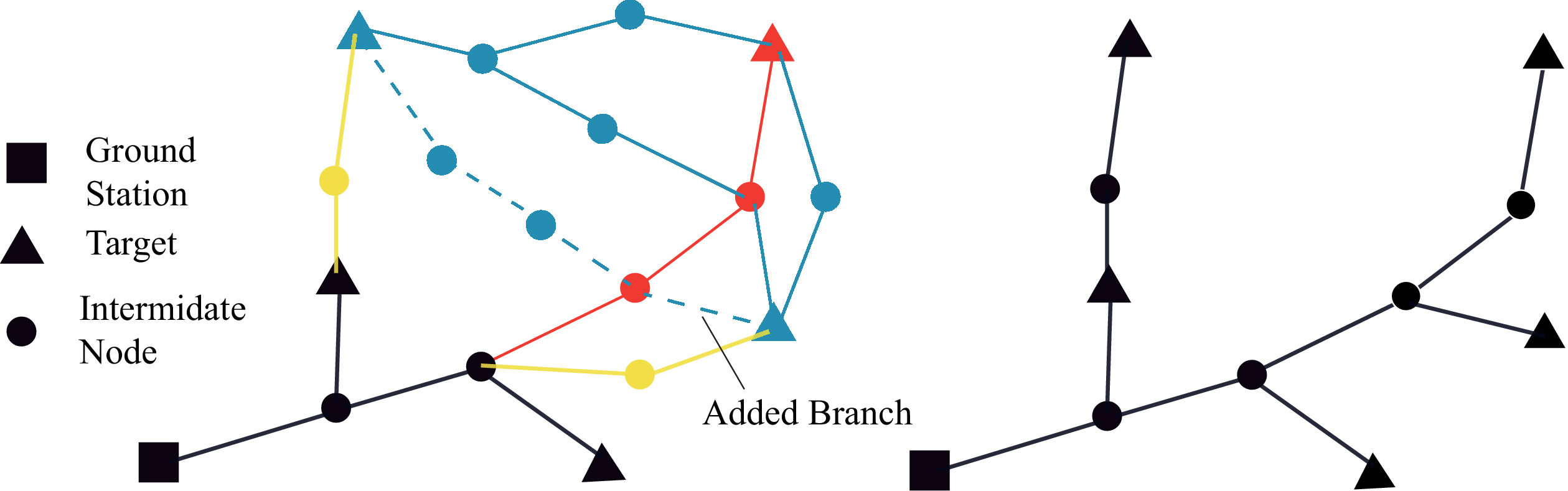}
	\caption{\textbf{Left}: The procedure of attaching new branches
		to the existing spanning tree.
		The targets in $\mathcal{I}$ (blue triangles) are added incrementally,
		with  nodes in $\mathcal{V}_\text{new}$ (red circles and triangle), current paths  $\Gamma_t^i$ (yellow lines) and best paths $\Gamma^\star$
		(dashed blue lines);
		\textbf{Right}: the derived communication network~$\mathcal{T}$.}
	\label{tree-span}
\end{figure}

As illustrated in Fig.~\ref{tree-span} and summarized in Alg.~\ref{AL:span-tree},
an iterative algorithm is proposed
to construct this tree~$\mathcal{T}$ incrementally.
It mainly consists of three steps:
(i) First, $\mathcal{T}$ is initialized,
along with the set of newly-added vertices~$\mathcal{V}_\text{new}$,
the set of indices $\mathcal{I}$ associated with the target positions,
and the temporary path $\Gamma_t^i$ for each target~$i \in \mathcal{I}$.
(ii) For each target $i\in \mathcal{I}$,
the set of paths $\widehat{\Gamma}$ from this target to every nodes in $\mathcal{V}_\text{new}$
are derived via the minimum-edge RRT$^\star$ in Line~\ref{algline:path-list}.
Then, the path with the minimum edge~$\Gamma^\star$ is selected from the $\widehat{\Gamma}$
in Line~\ref{algline:get-best-path}.
Particularly, if two paths have the same edges,
the one with a smaller cost from the root to the associated
destination node in~$\mathcal{V}_\text{new}$ is preferred.
Afterwards, the current-best path $\Gamma_t^i$ for target~$i$ is updated by
comparing it with~$\Gamma^\star$ in Line~\ref{algline:path-update}.
(iii) The optimal path~$\Gamma_t^{i^\star}$ of target~$i^\star$ with the minimum length
is found in Line~\ref{algline:best-index}.
Accordingly, the associated target~$i^\star$ is removed from the set of potential targets~$\mathcal{I}$.
As a result, the nodes extracted from~$\Gamma_t^{i^\star}$ are added to
the spanning tree as $\mathcal{V}_\text{new}$ in Line~\ref{algline:new-point} and~\ref{algline:add-branch},
along with their preceding nodes as parents.
This procedure is repeated until~$\mathcal{I}$ is empty,
meaning that all targets have been connected to this spanning tree~$\mathcal{T}$.
In other words, the tree greedily spans out a branch with the minimum edge in each iteration.

\begin{algorithm}[t!] \label{AL:span-tree}
	\caption{Optimize Communication Network \texttt{OptTree}()}
	\SetKwInOut{Input}{Input}
	\SetKwInOut{Output}{Output}
	\Input{$p_\text{g}$, $\mathcal{P}_\text{target}$, $\mathcal{O}$}
	\Output{$\mathcal{T}$, $\{\Gamma^i\}$}
	$\mathcal{T} \leftarrow \{ V_0 \}$;
	$\mathcal{V}_\text{new} \leftarrow \left\{ V_0 \right\}$;
	$\mathcal{I}\leftarrow \mathcal{M}$;
	$\Gamma_t^i \leftarrow \emptyset, \; \forall i \in \mathcal{I}$\; \label{algline:initial-path_tree}
	\While{$\mathcal{I} \neq \emptyset$}{
		\For{ $i$ \rm{\textbf{in}}  $\mathcal{I}$ parallel \label{algline:span} }{
			$\widehat{\Gamma} \leftarrow \texttt{MiniEdgeRRT}^\star(p_{\text{target},i},\, \mathcal{V}_\text{new},\, \mathcal{O})$ \label{algline:path-list}\;
			$\Gamma^\star \leftarrow$ Best path in $\widehat{\Gamma}$ \label{algline:get-best-path}\;
			$\Gamma_t^i \leftarrow \texttt{Compare}(\Gamma^\star,\, \Gamma_t^i)$ \label{algline:path-update} \;
		}
		$i^\star \leftarrow$ Index of the best $\Gamma_t^i$\label{algline:best-index} \;
		$\mathcal{I} \leftarrow \mathcal{I} \backslash i^\star$ \label{algline:remove} \;
		$\mathcal{V}_\text{new} \leftarrow$ Extract nodes from $\Gamma_t^{i^\star}$ \label{algline:new-point} \;
		$\mathcal{T} \leftarrow \mathcal{T} \cup \mathcal{V}_\text{new}$ \label{algline:add-branch} \;
	}
	Compute $\Gamma^i$, $i \in \mathcal{N}_s$ from $\mathcal{T}$ \label{algline:get-path} \;
\end{algorithm}

Given the spanning tree~$\mathcal{T}$ derived above,
the agents that are deployed to the target nodes are called \emph{searchers},
denoted by~$\mathcal{N}_s\subseteq \mathcal{N}$;
and the agents as the intermediate nodes are \emph{connectors},
denoted by~$\mathcal{N}_c\subseteq \mathcal{N}$.
Thus, the total number of agents required to form the tree~$\mathcal{T}$ is given by
$N=|\mathcal{T}|-1$.
Furthermore, the reference path for each searcher~$i\in \mathcal{N}_s$ is given
by its path~$\Gamma^i$ from the root to its associated target $p_{\text{target},n_i}$,
i.e., as a sequence of waypoints denoted
by~$\Gamma^i= \left\{ p_g,\cdots, p_{\text{target},n_i} \right\}$,
$\forall i\in \mathcal{N}_s$.
In contrast, the connectors do not have reference paths and serve as intermediate relays only.

\begin{remark}
	  Related works propose self-organizing topologies as lines \cite{Varadharajan2020} or
          minimum spanning trees (MST) \cite{Luo2020,Boldrer2021}.
	  Although they exhibit a high flexibility for dynamic tasks,
          they do not minimize the number of agents required for the deployment task.
	\hfill $\blacksquare$
\end{remark}

Computational complexity of Alg.~\ref{AL:span-tree} is similar to that of the traditional RRT$^{\star}$,
i.e., $n_\text{sample} \log(n_\text{sample})$ where $n_\text{sample}$ is the number of sampled nodes.
In our implementation, the planning time for each call to the minimum-edge RRT$^\star$ is bounded by a limited duration~$t_d$.
The parallel evaluation for each target is achieved via multi-threading,
thus the overall complexity is bounded by~$\mathcal{O}\big(M n_\text{sample} \log(n_\text{sample})\big)$.

\subsection{Collaborative Trajectory Planning via Distributed MPC}
Given the desired communication network~$\mathcal{T}$,
this section presents a distributed control strategy to maintain the designed
topology during collaborative motion, while avoiding collisions between agents and obstacles.
The core idea is to re-formulate this problem under the distributed MPC framework,
where the constraints are approximated as linear or quadratic inequalities.

To begin with,
denote by~$h >0$ the sampling time and $K \in \mathbb{Z}^+$ the planning horizon.
The planned state and control input at time $t+kh$ are defined
as $x^i_k(t)=[p^i_k(t),\, v^i_k(t)]$ and $u^i_k(t)$ of agent~$i$, respectively,
$\forall k\in \mathcal{K}= \{1,\cdots,K\}$.
The planned trajectory of agent $i$ at the time~$t$ is defined as
$\mathcal{P}^i=\left\{p^i_1, p^i_2, \cdots,p^i_K  \right\}$.
In the sequel, the time index ``$t$" is omitted wherever no ambiguity is caused.
Moreover, the so-called \emph{predetermined} trajectory is denoted by
$\overline{\mathcal{P}}^i(t)= \left\{ \overline{p}_{1}^{i}(t), \overline{p}_{2}^{i}(t), \cdots, \overline{p}_{K}^{i}(t) \right\}$,
where $\overline{p}_{k}^{i}(t) = p^i_{k+1}(t-h)$,
$k \in \{1,2,\cdots,K-1\}$ and $\overline{p}^i_{K}(t) = p^i_{K}(t-h)$.
Both trajectories should comply with the sampled dynamics of equation~\eqref{eq:problem}, i.e.,
\begin{equation} \label{dynamic-constraint}
	x_{k}^{i}=\mathbf{A} x_{k-1}^{i}+\mathbf{B} u_{k-1}^{i},
\end{equation}
where
$\mathbf{A}
=
\left[
\begin{array}{ccc}
	\mathbf{I}_3 & h \mathbf{I}_3  \\
	\mathbf{0}_3 & \mathbf{I}_3 \\
\end{array}
\right]$,
$\mathbf{B}
=
\left[
\begin{array}{ccc}
	\frac{h^2}{2} \mathbf{I}_3   \\
	h \mathbf{I}_3
\end{array}
\right]$;
and the constraints for velocities and inputs are the same as in equation~\eqref{eq:problem}.

\begin{figure}[t!]
	\centering
	\includegraphics[width=0.65\linewidth]{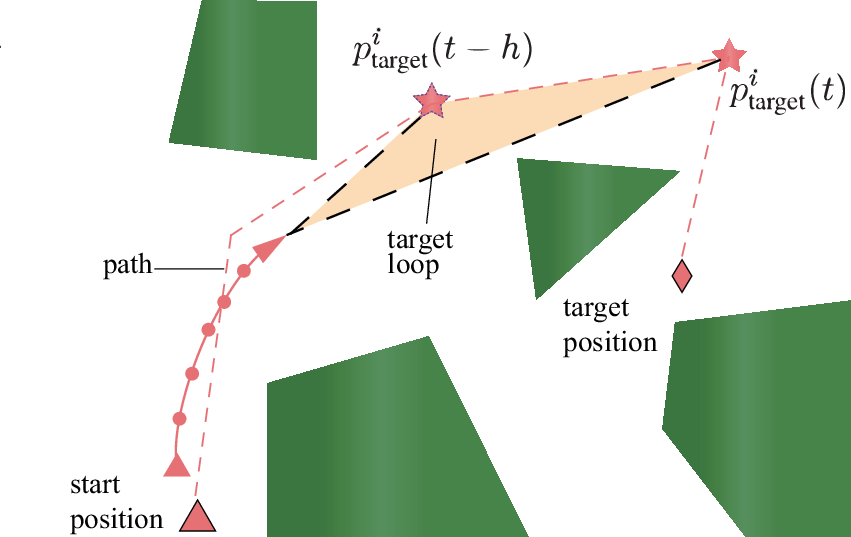}
	\caption{
		Illustration of how the intermediate target point is determined for a searcher~$i\in \mathcal{N}_s$.}
	\vspace{-0.15in}
	\label{get_target_point}
\end{figure}

\subsubsection{Collision Avoidance}
The collision avoidance among the agents can be re-formulated via the
modified buffered Voronoi cell (MBVC)  as proposed in our previous work~\cite{Chen2022-1}, i.e.,
\begin{equation} \label{eq:inter-constraint}
	{a_{k}^{i j}}^\mathrm{T} p_{k}^{i} \geq b_{k}^{i j}, \ \forall j\neq i, \;  \forall k \in \mathcal{K},
\end{equation}
where the coefficients are given by
$a_{k}^{i j}=\frac{ \overline{p}_{k}^{i}-\overline{p}_{k}^{j} } { \|\overline{p}_{k}^{i}-\overline{p}_{k}^{j}\|_{2} }$,
$b_{k}^{i j}=a_{k}^{i j^\mathrm{T}} \frac{\overline{p}_{k}^{i} + \overline{p}_{k}^{j}}{2}+\frac{r_{\min }^{\prime}}{2}$;
$ r^{\prime}_{\text{min}} = \sqrt{4{r_a}^{2}+h^{2} v_{\text{max} }^{2}} $.
On the other hand,
the collision avoidance between agent~$i\in \mathcal{N}$ and all obstacles in~$\mathcal{O}$
is realized by restricting its planned trajectory in a safe corridor formed by convex polyhedra.
The corridor serves as the boundary that separates the planned positions $p^i_k$
and the inflated obstacles given by
$\tilde{\mathcal{O}}(r_a) = \{p \in \mathcal{W} \ | \ \|p-p_o\|_2\leq r_a,\, p_o \in \partial \mathcal{O}\}$.
Specifically, the constraint over the planned trajectory is written as a linear
inequality:
\begin{equation} \label{eq:obstacle-constraint}
	{a_{k}^{i,o}}^\mathrm{T} p_{k}^{i} \geq b_{k}^{i,o},\, \forall k \in \mathcal{K},
\end{equation}
where $a_{k}^{i,o}$ and $b_{k}^{i,o}$ represents the linear boundary
for agent~$i$ at horizon $k$.
The detailed derivation is omitted here for brevity and refer
the readers to~\cite{Chen2022-2}.

Moreover,
an intermediate target point denoted by~$p_\text{target}^i$ is computed dynamically for each searcher~$i \in \mathcal{N}_s$.
Initially at time~$0$, $p_\text{target}^i = p_g$.
For time $t>0$, $p_\text{target}^i(t)$ is updated as the node on $\Gamma^i$
that is closest in distance to $p^i_{\text{target},n_i}$ and satisfies:
\begin{equation*}
	\textbf{Conv}\big(  \left\{  \Gamma^i\left( p_\text{target}^i(t-h), p_\text{target}^i(t)\right),
	\,\overline{p}^i_K(t) \right\}  \big) \cap \tilde{\mathcal{O}} = \emptyset,
\end{equation*}
where
$\Gamma^i( p_1, p_2 )$
denotes the sequence of nodes on $\Gamma^i$ between position $p_1$ to $p_2$;
and
$\textbf{Conv} (P) = \{\sum_{i=1}^n \theta_i p_i \,|\, \sum_i \theta_i =1,\, p_i \in P,\, \theta_i \geq 0, \, i=1,\cdots,n\}$
is the convex hull formed by points within the set~$P$.
For connector $i \in \mathcal{N}_c$, its intermediate target point
is determined by $\overline{p}^i_K$.
Then, the extended predetermined trajectory (EPT) is defined as
$\tilde{\mathcal{P}}^i= \left\{ \overline{p}_{1}^{i},\overline{p}_{2}^{i},\cdots,\overline{p}_{K+1}^{i} \right\} $
where
$\overline{p}_{K+1}^{i} = p^i_{\text{target}}$, $\forall i \in \mathcal{N}$.

% The connectivity maintenance between agent $i$ and $j$, $i>j$, $(i,j) \in E$,
% is realized via imposing constraints to their respective trajectory optimization.
% Before trajectory optimization,
% agent $i$ is assigned to formulate these constraints and share it with agent $j$.
% In particular, for agent $i$ that directly connects with ground station, it regards the station as agent $0$ with the extended predetermined trajectory
% $\tilde{\mathcal{P}}^0= \left\{ p_g, \cdots, p_g \right\}$.

\subsubsection{Connectivity Maintenance} \label{sub: connection-maintain}
Maintaining the desired topology during motion is the most challenging constraint
to deal with.
This constraint is decomposed into two sub-constraints:
(i) the relative distance between neighboring agents should be less than~$d_{c}$;
(ii) the LOS connecting neighboring agents should not be obstructed by obstacles.

The first sub-constraint is reformulated as a restriction that
each agent stays within a sphere of which
the center depends on its relative distance to the neighbors.
Specifically,
the position of agent~$i$ at horizon~$k$ is restricted by:
\begin{equation} \label{eq:circle-constraint}
	\left \| p_{k}^{i} - c^{i j}_k \right \| _2 \leq d_c, \, \forall j \in \mathcal{N}^i, \, \forall k \in \mathcal{K},
\end{equation}
where the center $c_k^{ij}\in \mathcal{W}$ is determined based on the EPT of agents $i$ and $j$
by considering three cases:
% Consider that $\overline{p}^i_k$, $\overline{p}^j_k$, $\overline{p}^i_{k+1}$ and $\overline{p}^j_{k+1}$ are taken into account.
% The determination of the center has following threefold:
%  inter-agent distance.
%\begin{figure}[t]
%	\centering
%	\includegraphics[width=0.9\linewidth]{figs/get_circle}
%	\caption{ The method that chooses the center of constrained circle where the square denotes $c^i_{k,\text{center}}$.  \textbf{Left}: Short inter-agent distance. \textbf{Medium}: long distance. \textbf{Right}: medium distance. }
%	\label{get_circle}
% \end{figure}
(i) If $ \| \overline{p}^i_k - \overline{p}^j_k \|_2 > d_w$
with $d_{w}>0$ being a chosen distance close to $d_{c}$,
$c_k^{ij}$ is set to $\frac{1}{2} (\overline{p}^i_k + \overline{p}^j_k)$;
(ii) If $\|p-c^i_{k,\text{center}}\|<d_{w}$ holds for all
$p\in \{\overline{p}^i_k, \overline{p}^j_k, \overline{p}^i_{k+1}, \overline{p}^j_{k+1}\}$
where $c^i_{k,\text{center}} = {\frac{1}{4} (\overline{p}^i_k+ \overline{p}^j_k+ \overline{p}^i_{k+1}+ \overline{p}^j_{k+1})}$,
it indicates that agents~$i$ and $j$ are in close distance and
$c_k^{ij}=c^i_{k,\text{center}}$;
(iii) Otherwise, they are in medium distance for which $c^{ij}_k$ is given by:
\begin{equation*}
	c_k^{ij} =  \eta^\star \frac{1}{2} (\overline{p}^i_k + \overline{p}^j_k)  + (1-\eta^\star) c^i_{k,\text{center}},
\end{equation*}
where $\eta^\star$ is derived by solving the following optimization:
\begin{equation} \label{eq:eta}
	\begin{aligned}
		& \eta^\star={\min}_{0\leq \eta \leq 1} \eta, \\
		\textbf{s.t.} \ & \| \overline{p}^i_k - \overline{c}_k^{ij} \|_2 \leq  \frac{d_w}{2},
		\| \overline{p}^j_k - \overline{c}_k^{ij} \|_2 \leq  \frac{d_w}{2},\\
		& \overline{c}_k^{ij} = \eta \frac{1}{2} (\overline{p}^i_k + \overline{p}^j_k) +  (1-\eta) c^i_{k,\text{center}},\\
	\end{aligned}
\end{equation}
which can be solved by the standard algorithm of bisection method \cite{Burden2015} with the search interval $[0,1]$.
Thus, the center and the allowed space are pushed towards the planned position in the next horizon,
yielding a more efficient trajectory.

\begin{figure}[t!]
	\centering
	\includegraphics[width=0.92\linewidth]{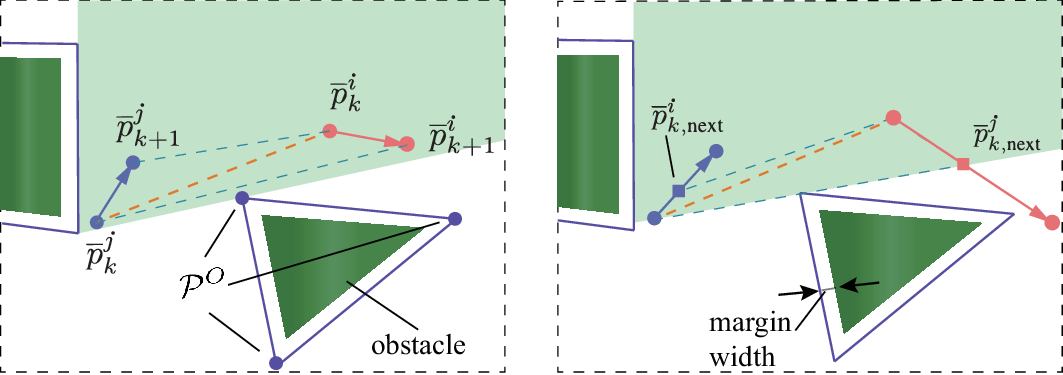}
	\caption{
		Illustration of the safe zone construction regarding the LOS constraint.
		The light green region splits the safe zone while the region with purple lines behalf the boundary of $\tilde{\mathcal{O}}$. \textbf{Left}: $\eta = 1$. \textbf{Right}: $\eta < 1$.
	}
	\vspace{-0.1in}
	\label{fig:get-safe-zone}
\end{figure}

Regarding the second LOS sub-constraint,
additional linear constraints are added to form a polyhedron safe zone for $p^i_k$ and $p^j_k$,
as illustrated in Fig.~\ref{fig:get-safe-zone}.
Before stating these constraints, two intermediate waypoints are computed as follows:
\begin{equation} \label{eq:p-next}
	\begin{aligned}
		\overline{p}^i_{k,\text{next}} = \overline{p}^i_{k} + \eta^\star (\overline{p}^i_{k+1}-\overline{p}^i_{k});\\
		\overline{p}^j_{k,\text{next}} = \overline{p}^j_{k} + \eta^\star (\overline{p}^j_{k+1}-\overline{p}^j_{k}),
	\end{aligned}
\end{equation}
where~$\eta^\star$ is the minimum~$\eta\in [0,\,1]$ such that
$\textbf{Conv} \big( \{ \overline{p}^i_{k}, \overline{p}^j_{k}, \eta \overline{p}^i_{k} + (1-\eta) \overline{p}^i_{k+1},  \eta \overline{p}^j_{k+1} + (1-\eta) \overline{p}^j_{k}\}   \big) \cap \ \hat{\mathcal{O}}(d_m) = \emptyset$,
where $\hat{\mathcal{O}}(d_m)$ is the set of inflated obstacles by a small margin~$d_m>0$.
Similar to optimization~\eqref{eq:eta}, the optimal~$\eta^\star$ is derived by the bisection method.
Given $\overline{p}^i_{k,\text{next}}$ and $\overline{p}^j_{k,\text{next}}$,
the linear constraints that separate the free space 
$\textbf{Conv}( \mathcal{P}^\text{free} )$
where 
$\mathcal{P}^\text{free} = \{ \overline{p}^i_k,\overline{p}^j_k,\overline{p}^i_{k,\text{next}},\overline{p}^j_{k,\text{next}} \}$
and any obstacle $O\in \mathcal{O}$ are obtained as follows:
\begin{equation} \label{eq:safe-zone-SVM}
	\begin{aligned}
		&  \max_{ \{a_{k}^{ij,s} ,  b_{k}^{ij,s} , \delta\} }    \; \delta  ,   \\
		\textbf{s.t.} \ &  { a_{k}^{ij,s} }^\mathrm{T} p  \geq \delta + b_{k}^{ij,s}, \ p\in\mathcal{P}^\text{free} ; \\
		&  { a_{k}^{ij,s} }^\mathrm{T} p \leq  b_{k}^{ij,s}, \ \forall p \in \mathcal{P}^O ;  \\
		&  \| a_{k}^{i,c }\|_{2} = 1;
	\end{aligned}
\end{equation}
where~$a_{k}^{ij,s} \in \mathbb{R}^3$, $b_{k}^{ij,s} \in \mathbb{R}$ and $\delta \in \mathbb{R}$ are optimization variables;
$\mathcal{P}^O$ are vertices of obstacle $O$ inflated by margin~$d_m$.
Since $\textbf{Conv} (\mathcal{P}^\text{free}) \cap \hat{\mathcal{O}} = \emptyset$,
$\delta>0$ must hold due to the theorem of separating hyperplane~\cite{Boyd2004}.
As a result, the optimization~\eqref{eq:safe-zone-SVM} can be solved
as a quadratic program similar to \cite{Chen2022-2}.
Consequently, given these coefficients~$a_{k}^{ij,s}$ and $b_{k}^{ij,s}$,
the linear constraints for agents $i$ and $j$ regarding the LOS constraint are given by:
\begin{equation} \label{eq:safe-zone-constraint}
	{ a_{k}^{ij,s}}^\mathrm{T} p_{k}^{i} \geq  b_{k}^{ij,s}, \, j \in \mathcal{N}^i, \, \forall k \in \mathcal{K},
\end{equation}
which is the separating hyperplane for obstacle~$O$.
This procedure is repeated from the nearest obstacle to the farthest one,
if this obstacle is not excluded by the existing hyperplane.

\subsection{Overall Algorithm}
The objective of each searcher~$i \in \mathcal{N}_s$ is to minimize the weighted summation
between the distance to its target point and velocity, i.e.,
\begin{equation} \label{eq:C^i-searcher}
	C^i = \frac{1}{2} Q_K \|p_{K}^{i}-p_{\text {target}}^{i}\|_2^2 + \frac{1}{2} \sum_{k=1}^{K-1}Q_{k}\|p_{k+1}^{i}-p_{k}^{i}\|_2^2,
\end{equation}
where $Q_k>0$ are positive definite matrices, $\forall k \in \mathcal{K}$.
On the other hand, each connector minimizes their distance to an intermediate point $\tilde{p}^i_k$
determined by its neighbors, i.e.,
\begin{equation} \label{eq:C^i-connector}
	\begin{split}
		C^i &= \frac{1}{2} \sum_{k=1}^{K} \|p_k^i-\tilde{p}^i_k \|_2^2;\\
		\tilde{p}^i_k &= \frac{  \alpha_c \sum_{j \in \mathcal{N}^i_c}\overline{p}^j_k  + \alpha_p \sum_{j \in \mathcal{N}_p^i} \overline{p}^j_k }{ \sum_{j \in \mathcal{N}_c^i} \alpha_c  + \sum_{j \in \mathcal{N}_p^i} \alpha_p};
	\end{split}
\end{equation}
where $\alpha_c, \alpha_p>0$ are constant weighting parameters for the set of child
agents~$\mathcal{N}^i_c$ and parent agents~$\mathcal{N}^i_p$ within the tree~$\mathcal{T}$, respectively.
Namely, $\tilde{p}^i_k$ is the weighted average position of all parent and child agents,
where $\alpha_c$ is set larger than $\alpha_p$ to bias the child agents.
Thus, the overall optimization for the trajectory planning of agent~$i\in \mathcal{N}$
is summarized as follows:
\begin{equation} \label{convex-program}
	\begin{aligned}
		&\min _{ \mathbf{u}^i, \mathbf{x}^i }\; C^{i} \\
		\mathbf{s.t.} \quad &  \eqref{dynamic-constraint}, \eqref{eq:inter-constraint},
		\eqref{eq:obstacle-constraint}, \eqref{eq:circle-constraint},
		\eqref{eq:safe-zone-constraint}; \\
		& v^i_K=\mathbf{0}_3,
	\end{aligned}
\end{equation}
where $\mathbf{u}^i$ and $\mathbf{x}^i$ are the stacked vector of $u^i_{k-1}$ and $x^i_k$,
$\forall k \in \mathcal{K}$;
the last constraint is enforced to avoid a overaggressive final state.
It is worth noting that the optimization~\eqref{convex-program} is a quadratically constrained
quadratic program (QCQP) as it only has a quadratic objective function
in addition to linear and quadratic constraints.
Thus, it can be fast resolved by off-the-shelf convex optimization solvers.

\begin{algorithm}[t!] \label{AL:IMPC-MS}
	\caption{Overall Algorithm}\label{algorithm}	\SetKwInOut{Input}{Input}
	\SetKwInOut{Output}{Output}
	\Input{$p_g$, $\mathcal{P}_\text{target}$, $\mathcal{O}$}
	$\Gamma^i \leftarrow$ \texttt{OptTree}$(p_g,\mathcal{P}^i_\text{target},\mathcal{O})$ \label{algline:formulated-network}\;
	$\overline{\mathcal{P}}^i(t_0) = p^{i}(t_0)\cdots p^{i}(t_0)$ \label{algline:impc-init}\;
	\While{task not accomplished}
	{
		\For{$i \notin \mathcal{N}_a$ \rm{concurrently} }{

			$\overline{\mathcal{P}}^j (t)$ $\leftarrow$ Receive  via communication \label{algline:commu} \;

			$\psi^{i,j}$ $\leftarrow$ Derive constraints~\eqref{eq:circle-constraint} and \eqref{eq:safe-zone-constraint} \label{algline:sight-cons}\;

			$\psi^{i,l}$ $\leftarrow$ Receive from agent~$j$\label{algline:commu-cons} \;
			$\zeta^i \leftarrow \psi^{i,j} \cup \psi^{i,l}$ \label{algline:cons-collection}\;

			$\zeta^i \leftarrow$ Add constraints \eqref{eq:inter-constraint} and \eqref{eq:obstacle-constraint} \label{algline:impc-inter-cons}\;
			$\mathcal{P}^i(t)$ $\leftarrow$ Solve optimization~\eqref{convex-program} based on $\zeta^i$ \label{algline:convex-programming}\;
		}
		$t \leftarrow t+h$\;
	}
\end{algorithm}

The complete trajectory planning algorithm is summarized in Alg.~\ref{AL:IMPC-MS}.
To begin with, the optimization of communication network is performed via Alg.~\ref{AL:span-tree} (Line~\ref{algline:formulated-network}).
Afterwards, the predetermined trajectories of all agents are initialized.
In the main loop of execution,
the agents exchange their predetermined trajectories via communication (Line~\ref{algline:commu}).
At each step,
the connectivity preserving constraints w.r.t. the parent agent
$\psi^{ij} = \{ c^{ij}_k, a^{ij,s}_k,b^{ij,s}_k \}$ are calculated (Line~\ref{algline:sight-cons})
and exchanged among them (Line~\ref{algline:commu-cons}).
In combination with the constraints for collision avoidance,
the overall constraints for optimization~\eqref{convex-program} are summarized as $\zeta^i$
(Line~\ref{algline:impc-inter-cons}).
Given these constraints, the optimization~\eqref{convex-program} is formulated and solved locally
by each agent~$i\in \mathcal{N}$ to obtain its planned trajectory (Line~\ref{algline:convex-programming}).
It should be mentioned that the above procedure is applied to
all agents including the searchers and connectors.

\begin{theorem}  \label{pro:connectivity-guarantee}
	If the agents are initially collision-free and connected,
	the system remains collision-free and the communication graph $G(t)$ remains connected at all time.
\end{theorem}

\begin{proof} (Sketch)
	It can be proven that the overall optimization in~\eqref{convex-program}
	is recursively feasible, in a way similar to our previous work~\cite{Chen2022-2}.
	Consequently, all constraints enforced in~\eqref{convex-program} are satisfied,
	i.e., no collision among the agents, no collision between the agents and obstacles,
	neighboring agents in the communication network remain connected during motion.
\end{proof}

\section{Experiments}

This section presents the numerical simulations and hardware experiments to validate the proposed scheme.
The methods are implemented in Python3 and CVXOPT \cite{cvxopt} is chosen as the optimization solver.
All tests are performed on a computer with Intel Core i9 @3.2GHz.
Simulation and experiment videos can be found in the supplementary files.

\begin{figure}[t]
	\centering
	\includegraphics[width=0.95\linewidth]{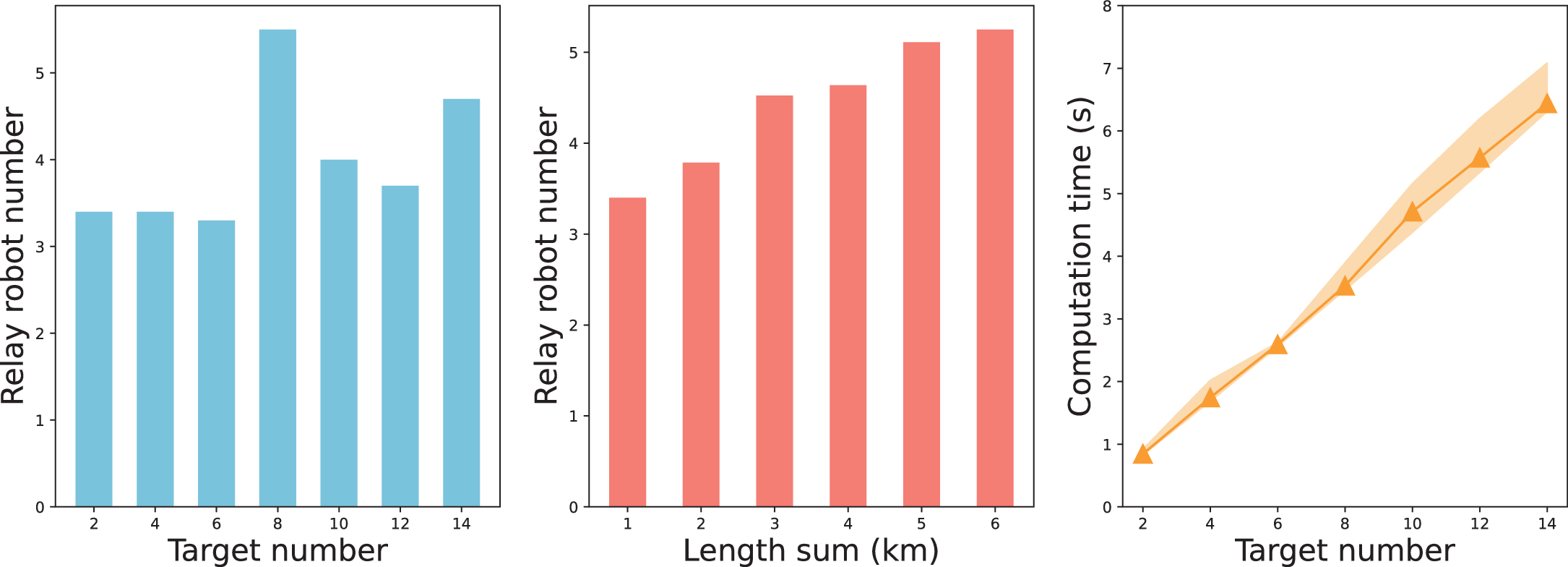}
	\caption{
		Comparisons of the communication network~$\mathcal{T}$ with different number of targets (\textbf{Left}) and the summed length of the shortest paths to all targets (\textbf{Middle}).
		And the computation time with different target number (\textbf{Right}).
	}
	\vspace{-0.1in}
	\label{fig:network-evaluation}
\end{figure}

\subsection{Numerical Simulations}

To mimic the quadrotor used in the field implementations,
the maximum velocity and acceleration of all agents are set
to~$15 \rm{m/s}$ and $3 \rm{m/s^2}$, respectively.
The replanning interval for MPC is set to the sampling time~$h=0.5$s.
Additionally, the safety radius of an agent is set to $r_a=2$m,
while the communication range and the margin of safe zone are chosen as
$d_c = 150$m and $d_m = 3$m, respectively.

\subsubsection{Optimization of Communication Network}
To begin with, the proposed Alg.~\ref{AL:span-tree} is systematically evaluated
under different number of targets.
More specifically, consider the workspace shown in Fig.~\ref{fig:overall}
which has a cubic volume from $(0,0,0)$ to $(500, 500, 100)$
with cluttered and convex-shaped obstacles.
Different numbers of target points are randomly generated within the obstacle-free space,
for which the proposed Alg.~\ref{AL:span-tree} is applied to compute
the communication network.
The allowed duration of each call to \texttt{MiniEdgeRRT}$^\star$ is limited $0.4$s.
For each target number, the tests are repeated for $15$ times.
The results are summarized in Fig.~\ref{fig:network-evaluation}.
It can be seen that
via the proposed algorithm for topology optimization,
the number of required relay agents does not increase linearly with the number of targets.
The same pattern can also be found in the summed length of the shortest paths
from the ground station to each target in~$\mathcal{T}$ w.r.t. the number of relay agents.
Lastly, the overall computation time to derive~$\mathcal{T}$ is recorded,
which increases linearly with the number of targets.
This is consistent with our theoretical analysis that Alg.~\ref{AL:span-tree}
has complexity~$\mathcal{O}(M)$ with the number of targets~$M$.

\begin{table} [t]
	\caption{Comparison of Network Topology (Averaged over all Tests)}
	\label{table:comparison}
	\begin{tabular}{ccccc|cccc}
		\toprule
		Metric & \multicolumn{4}{c}{$N_r$}  & \multicolumn{4}{c}{$N_h$} \\
		\cmidrule(r){2-5} \cmidrule(r){6-9}
		Targets & 2 & 4 & 6 & 8 & 2 & 4 & 6 & 8\\
		\midrule
		MST & 6.5  & 8.3 & 8.5 & 8.7 & 61 & 75.5 & 76.2 & 86.2\\
		\textbf{Ours} & 6.3  & 7.7 & 8.4 & 8.4 & 56.5 & 66.8 & 64.3 & 76.8\\
		\bottomrule
	\end{tabular}
\end{table}

\begin{figure}[t]
	\centering
	\includegraphics[width=0.45\linewidth]{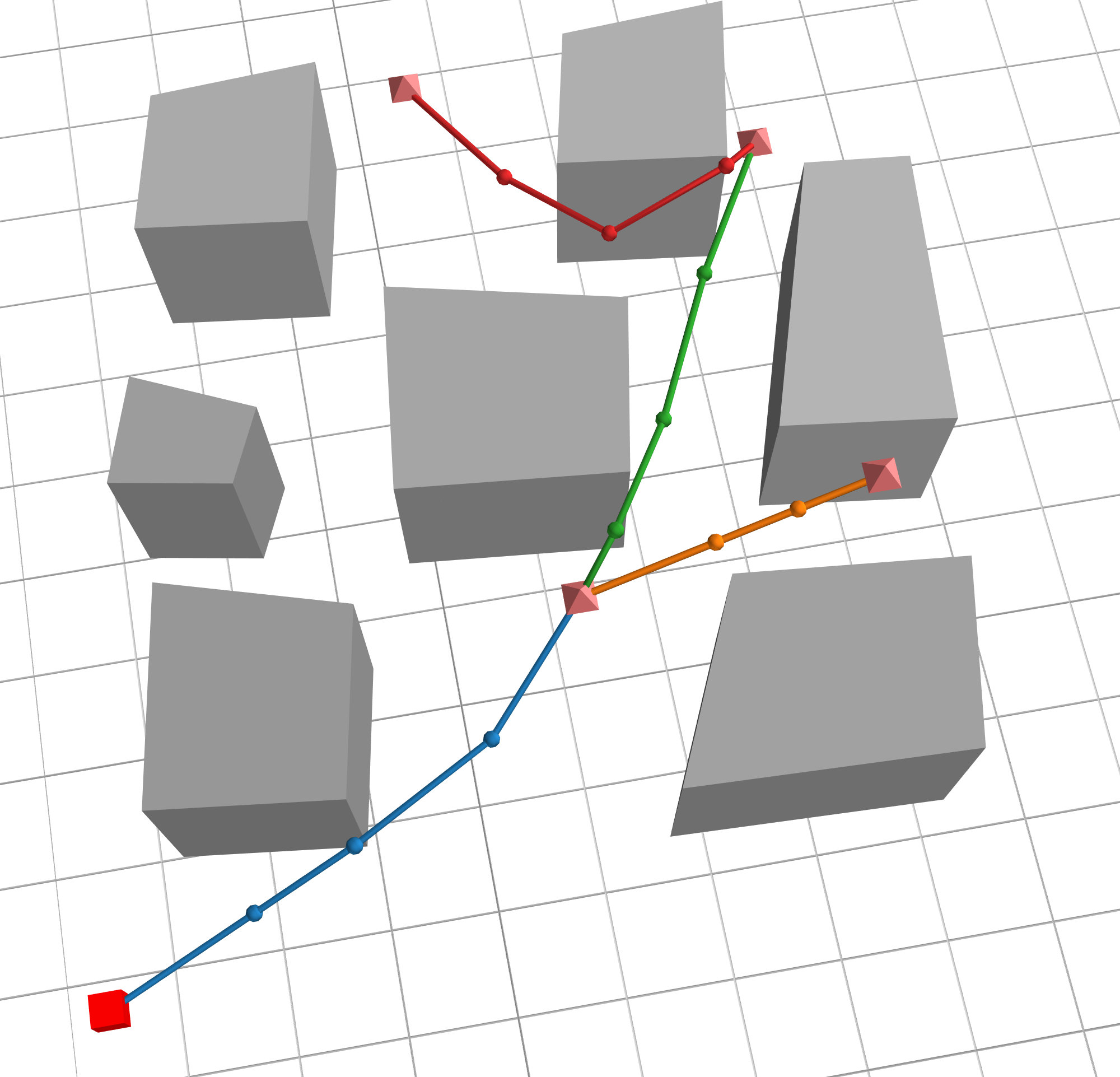}
	\includegraphics[width=0.45\linewidth]{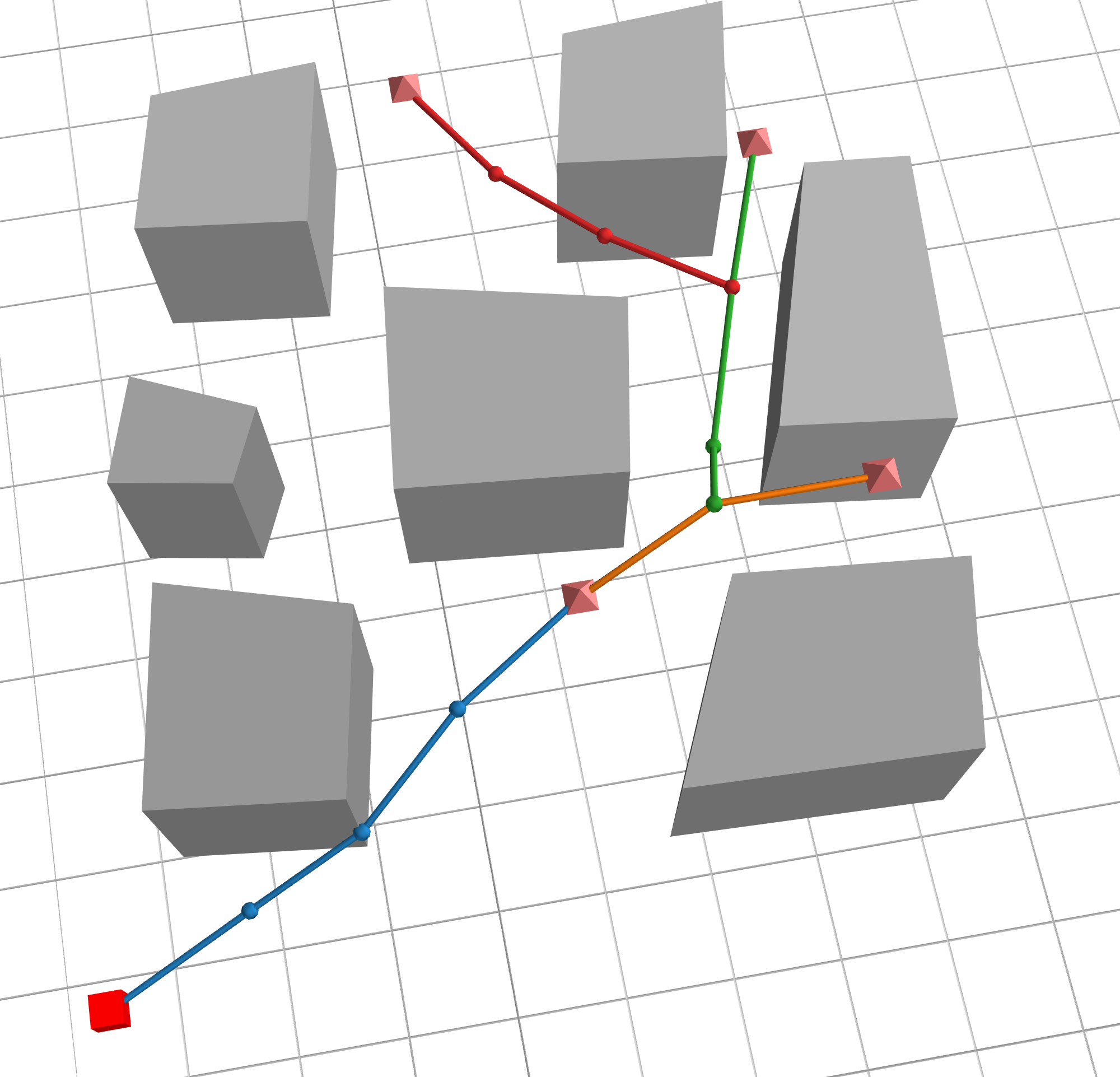}
	\caption{
          Comparison of the final topology.
          The MST formed by~\cite{Goodrich2015} requires $10$ relay agents (\textbf{Left}),
          while our method requires only $8$ (\textbf{Right}).
         }
     \vspace{-0.1in}
	\label{fig:network-compare}
\end{figure}

Furthermore, the proposed approach is compared against the Minimum Spanning Tree (MST)
from~\cite{Goodrich2015}, which is the predominant method to optimize a network topology.
For different number of targets, both methods are performed $10$ times and
the following two metrics are compared:
(i) the average number of required relay agents~$N_r$;
and (ii) the average of total number of hops from the ground station to all targets $N_h$.
The second metric is an important measure of the network reliability
as communication routing over longer paths has a higher error rate.
As summarized in Table~\ref{table:comparison},
the proposed method requires in average less relay agents as well as less hops of communication.
For instance, the total number of hops via MST under the case of $8$ targets is $12\%$ more than
our method.
A concrete example of $3$ targets is shown in Fig.~\ref{fig:network-compare},
where the MST requires $10$ relay agents while our method requires $8$ relay agents.
This is due to the fact that intermediate relay agents can be shared by the paths to different targets via our method.
Additionally, the number of hops is reflected by the number~$N_\text{root}+N_\text{segment}$ when choosing~$\Gamma^\star$ in the design of our algorithm, 
which induces a less $N_h$.

\subsubsection{Collaborative Motion Planning}

\begin{figure}[t]
	\centering
	\includegraphics[width=0.85\linewidth]{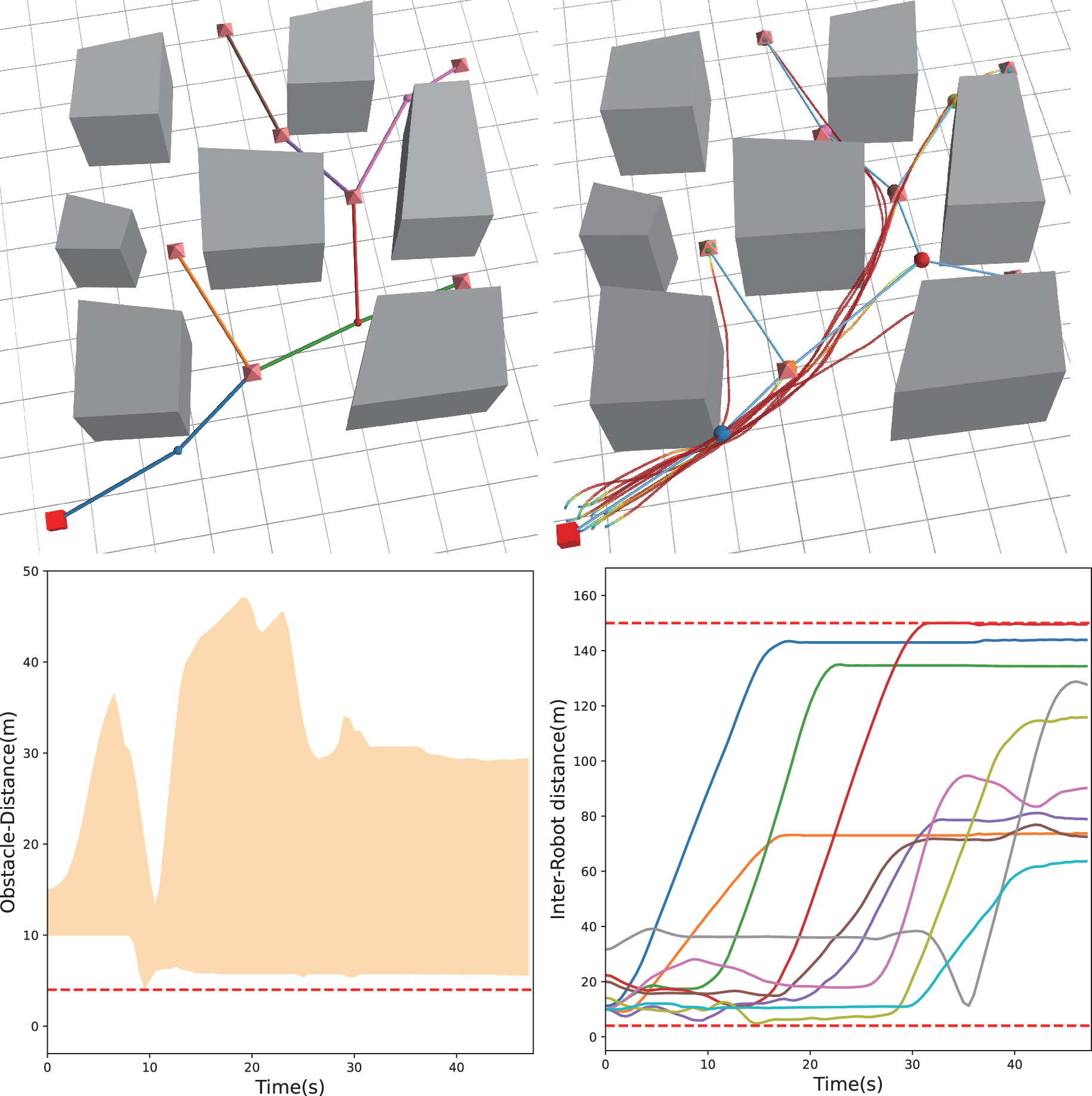}
	\caption{ \textbf{Top-Left}: the optimized network.
			  \textbf{Top-Right}: the trajectories of agents.
			  \textbf{Bottom-Left}: the distance between LOS and the closet obstacle.
			  \textbf{Bottom-Right}: the distance between interconnected agent.
			}
	\label{fig:simulation}
\end{figure}

Given the optimized network topology, the collaborative motion planning strategy
as described in Alg.~\ref{AL:IMPC-MS} is applied.
Regarding the design parameters,
the warning distance of communication, i.e., $d_w$ is set as $142$m while the margin of LOS is set as $3$m.
Meanwhile, the weighting parameters of the objective function for a connector
are chosen as $\alpha_c=3.0$ and $\alpha_p=1.0$.
The resulting trajectories under one scenario of~$7$ targets are visualized
in Fig.~\ref{fig:simulation},
along with two plots of the time-varying distances.
The complete mission is accomplished within $46$s.
It can be seen that all planned edges in~$\mathcal{T}$ are kept during motion since
the minimum distance between the LOS of any neighboring agents in $\mathcal{T}$
and the set of static obstacles is strictly larger than~$d_m=3$m over all time,
i.e., no obstruction with all LOS.
Furthermore, the resulting trajectories are safe as the minimum distance between
any pair of agents is also strictly positive over all time,
while the distance between neighboring agents is less than~$d_c=150$m over all time.
Note that the average speed of all $7$ searchers
reaches $80\%$ of the maximum speed yielding an efficient execution of
the desired topology, while ensuring safety and connectivity.
On the other hand,
the computation time of each agent during each iteration of the proposed MPC is recorded,
which is split into the time to derive the constraints related to the connectivity maintenance
in~\eqref{eq:safe-zone-SVM} and the time to solve the final constrained optimization in~\eqref{convex-program}.
Particularly, the average planning time per agent over the $100$ iterations is~$35$ms,
of which~$17.3$ms is used to derive the constraints in equation~\eqref{eq:safe-zone-SVM}
and $17.7$ms for solving optimization~\eqref{convex-program}.
However, the former exhibits a much larger variance due to different obstacle layouts across the trajectories.

\begin{figure}[t]
	\centering
	\includegraphics[width=0.95\linewidth]{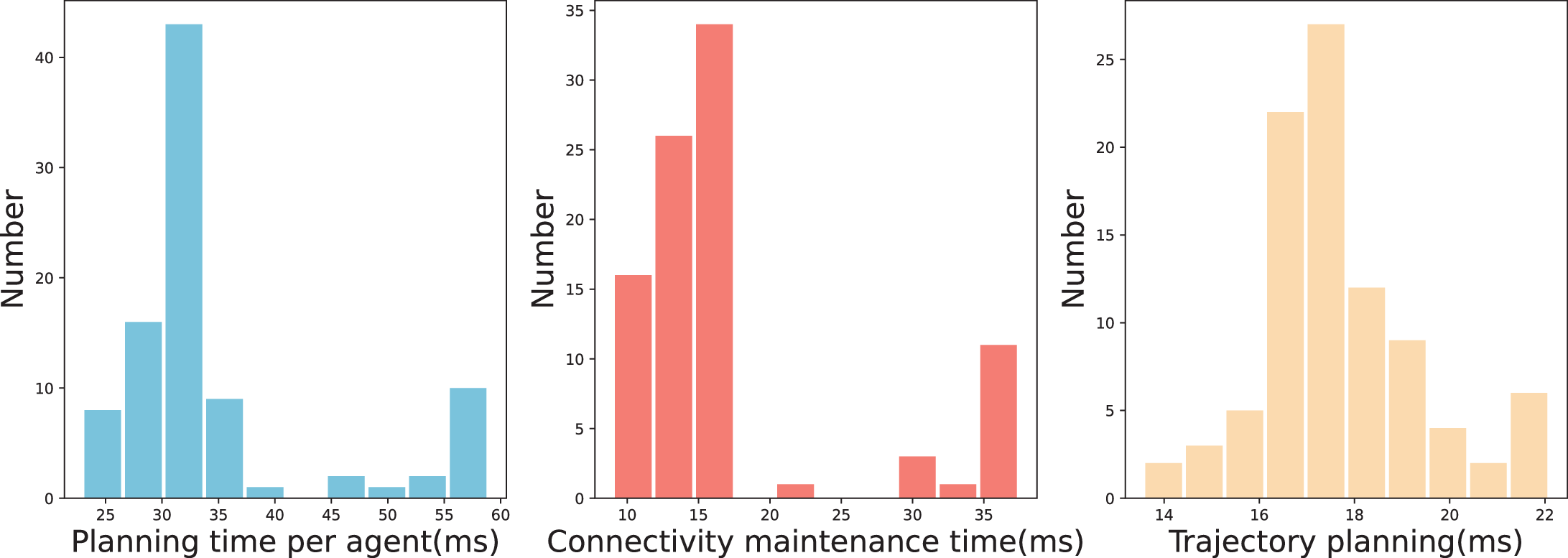}
	\caption{
          Distribution of the average computation time of each iteration
          for all agents (\textbf{Left}),
          the average time to derive the constraints related to the connectivity maintenance
          in~\eqref{eq:safe-zone-SVM} (\textbf{Middle}),
          and the average time to solve the final constrained optimization
          in~\eqref{convex-program} (\textbf{Right}).
  	}
  	\vspace{-0.15in}
	\label{fig:time-cost}
\end{figure}

\begin{figure}[t]
	\centering
	\includegraphics[width=0.45\linewidth]{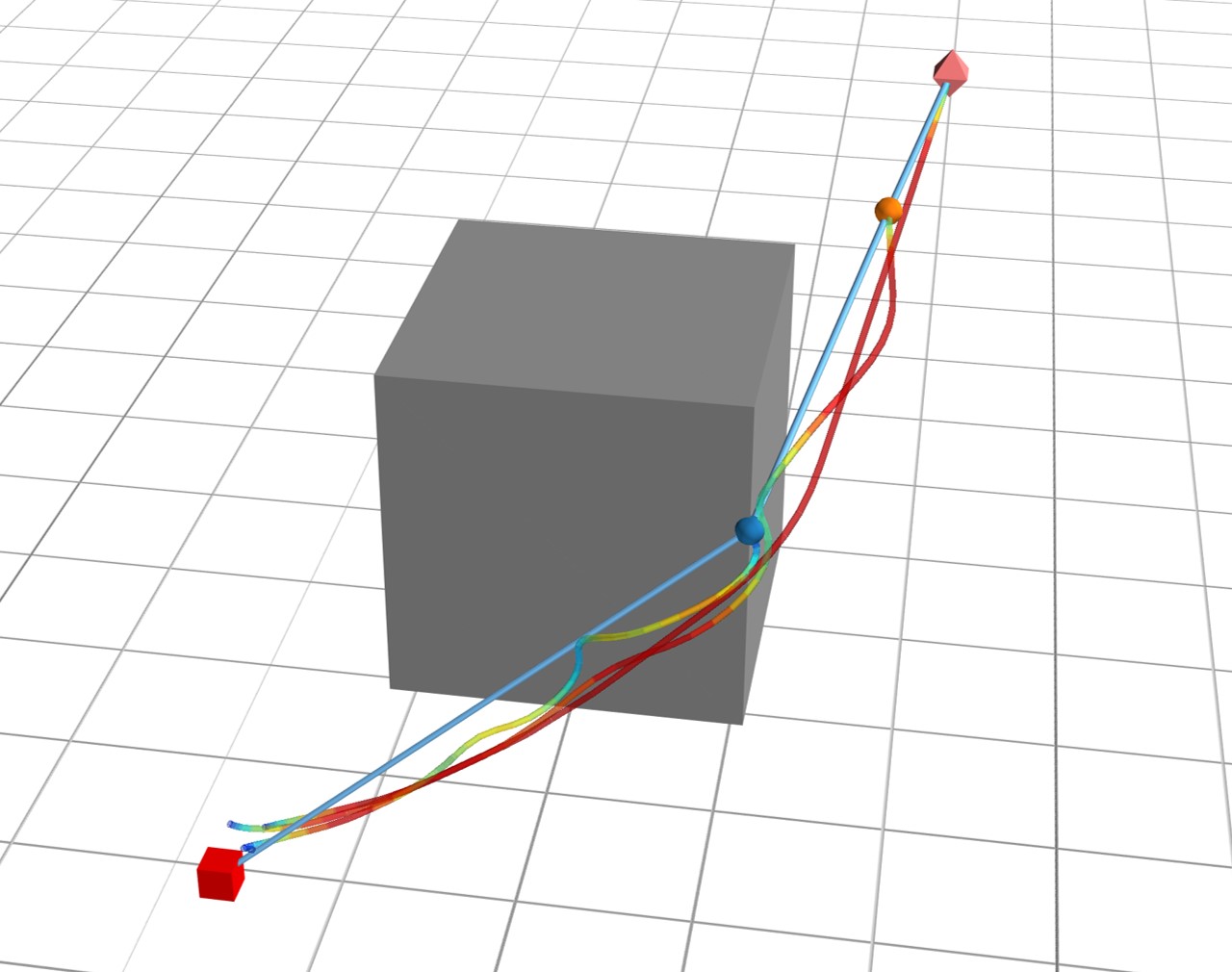}
	\includegraphics[width=0.45\linewidth]{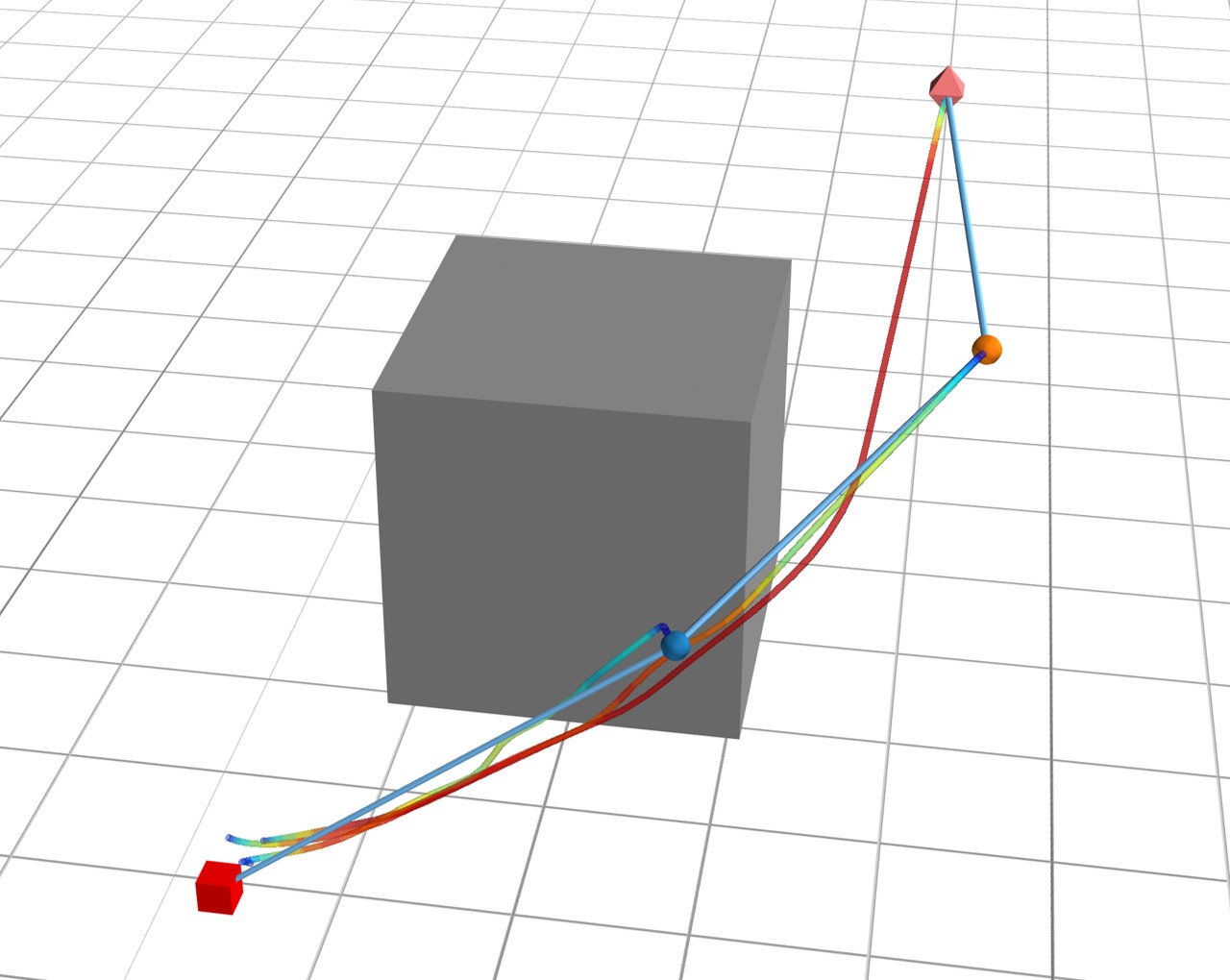}
	\quad
	\includegraphics[width=0.45\linewidth]{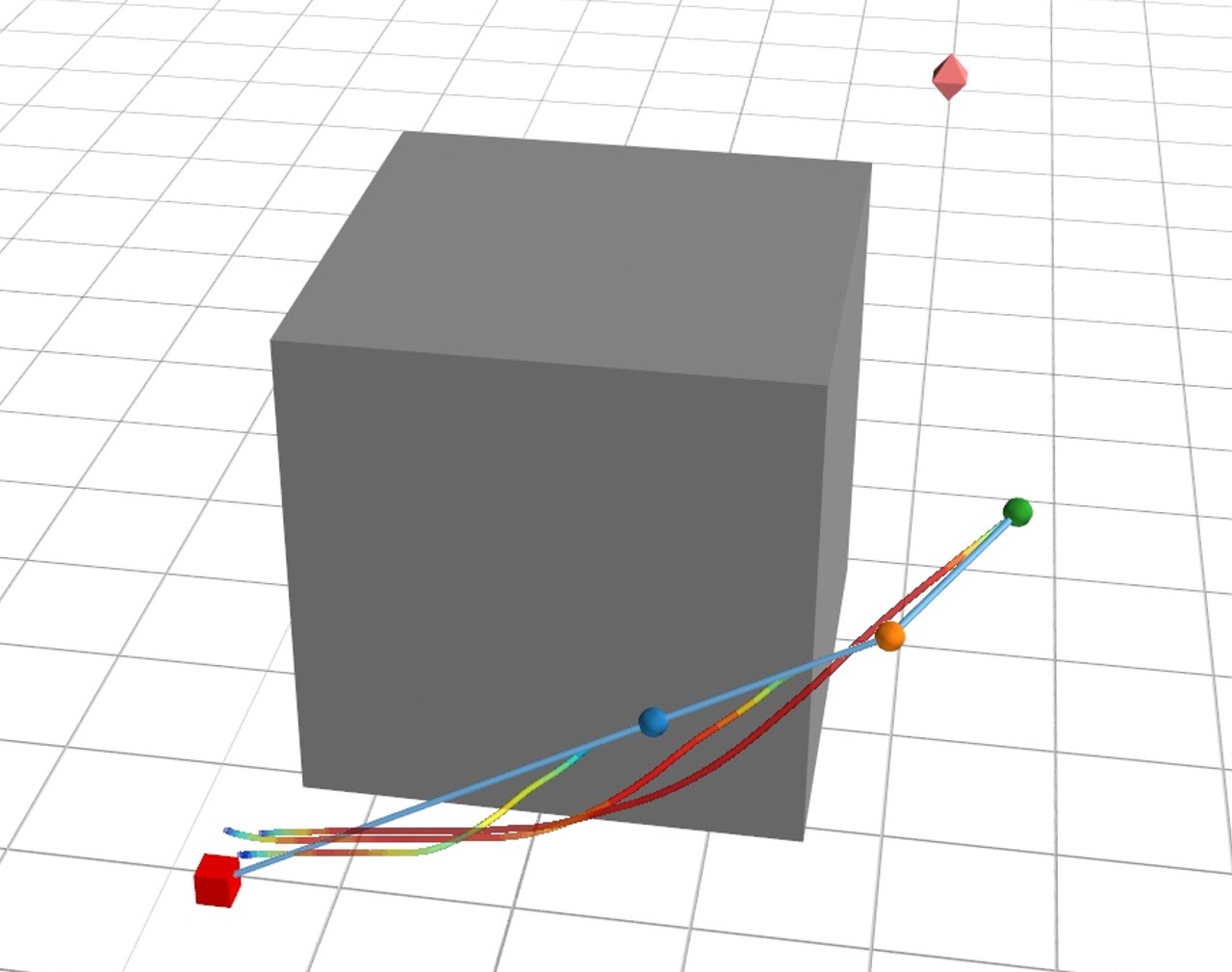}
	\includegraphics[width=0.45\linewidth]{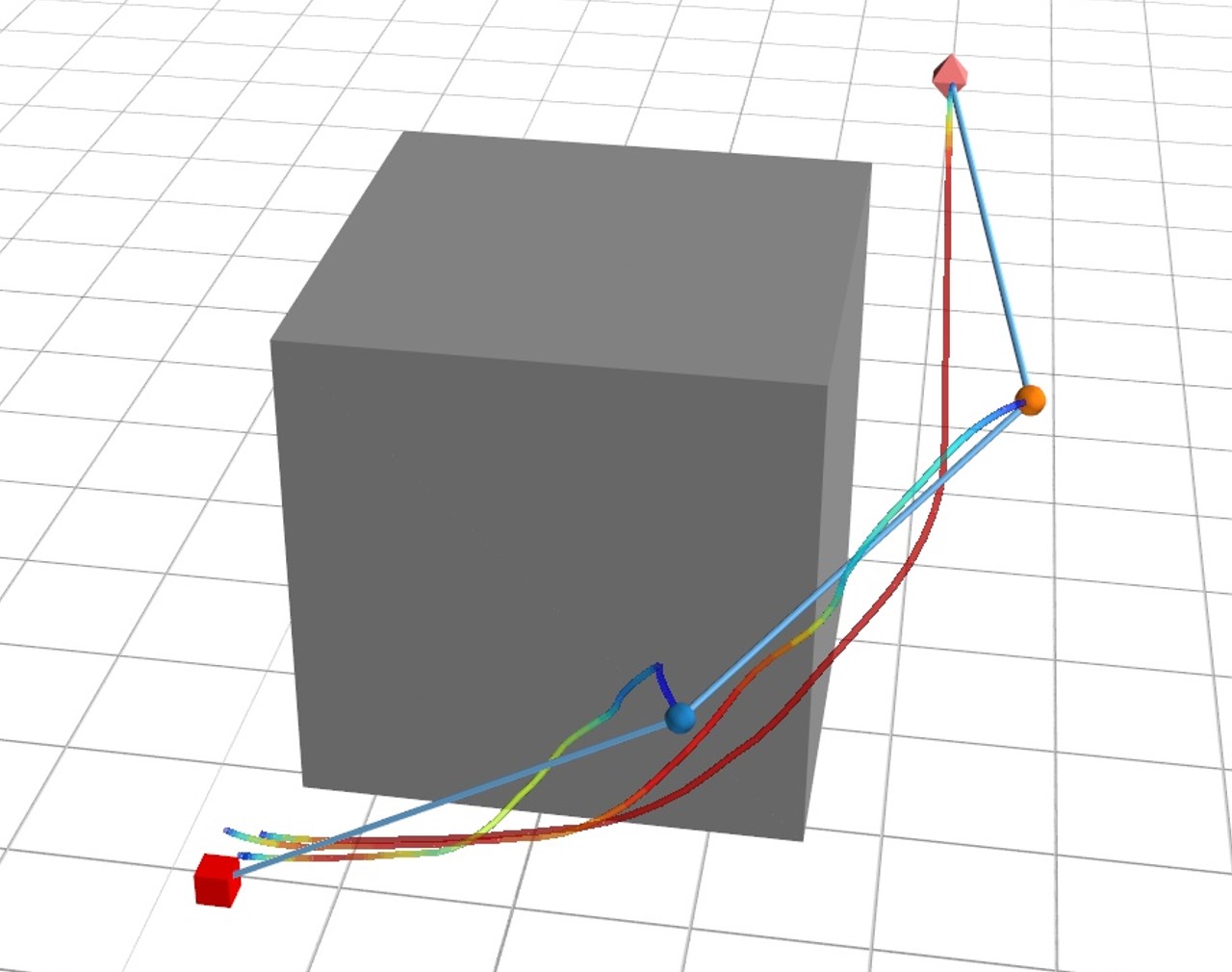}
	\caption{
	  Comparisons between the method in \cite{Caregnato2022} for a small and large obstacle~(\textbf{Left})
          and our method for the same workspace~(\textbf{Right}).}
	\label{fig:motion-compare}
\end{figure}

For comparison, the method proposed in~\cite{Caregnato2022} is implemented,
which handles the LOS constraint and the collision avoidance via mixed integer linear programs (MILP)
and solved via CVXPY \cite{cvxpy}.
More concretely, the obstacles are encoded as the union of numerous half spaces.
The results over different obstacles are shown in Fig.~\ref{fig:motion-compare}.
For a small cubic obstacle ($100 {\rm m}$ width),
it takes $857$ms for the method in~\cite{Caregnato2022} to
find a longer and more oscillatory trajectories of total length~$850$m
in comparison to ours method that takes merely $65$ms to plan trajectories of~$765$m.
However, given a larger cubic obstacle ($150 {\rm m} $ width), 
the method in \cite{Caregnato2022}
fails to generate a solution due to the large obstruction from the obstacle.
In contrast, our method can still accomplish the deployment task in $30$s
with the traveling distance of $843$m which validates its robustness
towards different environments.

\begin{figure}[t!]
	\centering
	\includegraphics[width=0.96\linewidth]{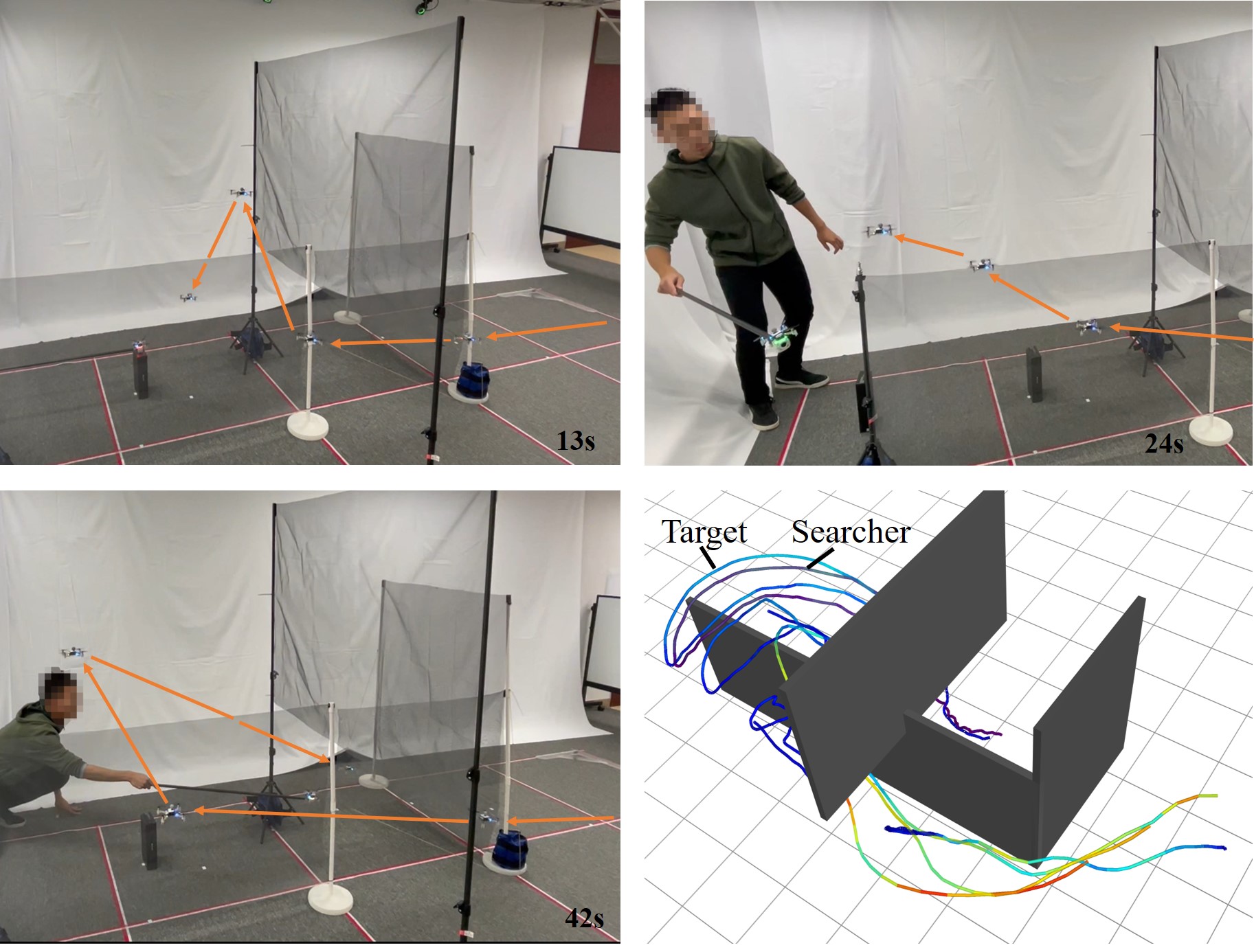}
	\caption{Adaptation of the network topology and the agent trajectories when the target is moving.}
	\vspace{-0.1in}
	\label{fig:realfly}
\end{figure}

\subsection{Hardware Experiment}

To further demonstrate the efficiency of our proposed method,
hardware experiments are performed on a fleet of nano-quadrotors based on Crazyswarm.
Each quadrotor is represented by an ellipsoid of diameter $0.24$m in the $xy$ plane and $0.6$m in the $z$ axis to concern the air turbulence in $z$ axis.
Moreover, its maximum velocity and acceleration is $1.0{\rm m/s}$ and $1.0{\rm m/s^2}$, respectively.
The sampling time~$h$ is set to $0.2s$ and the horizon length is chosen as $K=8$.
As shown in Fig.~\ref{fig:overall} and Fig.~\ref{fig:realfly}, the workspace consists of several 3D walls.
of which the size and position are fully-known.
The states of all quadrotors are tracked by the indoor motion capture system OptiTrack.
Alg.~\ref{AL:IMPC-MS} is performed on a central computer but multi-processing.

\begin{figure}
	\centering
	\includegraphics[width=0.98\linewidth]{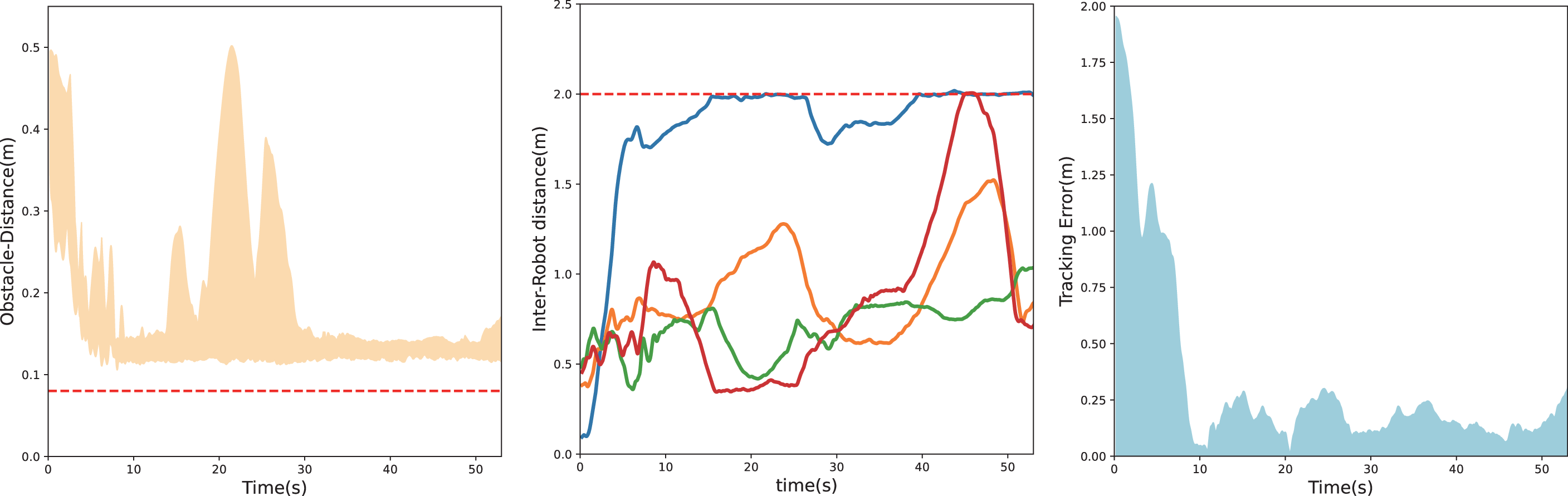}
	\caption{\textbf{Left}: the minimum distance between the LOS of any neighboring agents and the obstacles;
          \textbf{Middle}: the distance between any pair of neighboring agents.
          \textbf{Right}: the tracking error between the dynamic target and the associated searcher.}
	\label{fig:experiment}
	\vspace{-0.15in}
\end{figure}

For the first experiment,
two targets are set between the walls.
The derived communication topology by Alg.~\ref{AL:span-tree} is shown in Fig.~\ref{fig:overall},
which consists of $4$ relay agents and $2$ service agents in the ``F''-like shape.
The collaborative motion from the initial positions to the target positions takes
around~$10$s, during which the average velocity of all agents reaches $0.4{\rm m/s}$.
Both the collision avoidance and connectivity maintenance are ensured by their relative distances.
To further demonstrate the robustness and dynamic adaptation of the proposed method,
given a linear topology,
the target is moved manually as shown in Fig.~\ref{fig:realfly}.
Thus, the corresponding searcher needs to track the target
while the whole fleet moves accordingly to ensure that the network connectivity
is still maintained in motion.
As shown in Fig.~\ref{fig:experiment},
the distance between the LOS of any neighboring agents and any obstacle
is kept larger than $d_m=0.08$m at all time,
while the distance between any neighboring agents is almost kept below the communication radius $d_c=2.0$m.
It is worth noting that the tracking error between the searcher and the moving target
has been kept below $0.25$m at all time.

\section{Conclusion}

This work has presented a multi-UAV deployment method in obstacle-cluttered environments
where the LOS connectivity is maintained during motion at all time.
Different from most exiting work, the team size is optimized rather than given,
and the distributed MPC has a guarantee on the feasibility and safety.
Future work includes broader task applications such as coverage and surveillance.

\bibliographystyle{IEEEtran}
\bibliography{REF}

\begin{thebibliography}{10}
\providecommand{\url}[1]{#1}
\csname url@samestyle\endcsname
\providecommand{\newblock}{\relax}
\providecommand{\bibinfo}[2]{#2}
\providecommand{\BIBentrySTDinterwordspacing}{\spaceskip=0pt\relax}
\providecommand{\BIBentryALTinterwordstretchfactor}{4}
\providecommand{\BIBentryALTinterwordspacing}{\spaceskip=\fontdimen2\font plus
\BIBentryALTinterwordstretchfactor\fontdimen3\font minus
  \fontdimen4\font\relax}
\providecommand{\BIBforeignlanguage}[2]{{%
\expandafter\ifx\csname l@#1\endcsname\relax
\typeout{** WARNING: IEEEtran.bst: No hyphenation pattern has been}%
\typeout{** loaded for the language `#1'. Using the pattern for}%
\typeout{** the default language instead.}%
\else
\language=\csname l@#1\endcsname
\fi
#2}}
\providecommand{\BIBdecl}{\relax}
\BIBdecl

\bibitem{Chung2018}
S.-J. Chung, A.~A. Paranjape, P.~Dames, S.~Shen, and V.~Kumar, ``A survey on
  aerial swarm robotics,'' \emph{IEEE Transactions on Robotics}, vol.~34,
  no.~4, pp. 837--855, 2018.

\bibitem{Nouyan2009}
S.~Nouyan, R.~Gross, M.~Bonani, F.~Mondada, and M.~Dorigo, ``Teamwork in
  self-organized robot colonies,'' \emph{IEEE Transactions on Evolutionary
  Computation}, vol.~13, no.~4, pp. 695--711, 2009.

\bibitem{Dorigo2013}
M.~Dorigo, D.~Floreano, L.~M. Gambardella, F.~Mondada, S.~Nolfi, and e.~a.
  Baaboura, Tarek, ``Swarmanoid: A novel concept for the study of heterogeneous
  robotic swarms,'' \emph{IEEE Robotics and Automation Magazine}, vol.~20,
  no.~4, pp. 60--71, 2013.

\bibitem{Stephan2017}
J.~Stephan, J.~Fink, V.~Kumar, and A.~Ribeiro, ``Concurrent control of mobility
  and communication in multirobot systems,'' \emph{IEEE Transactions on
  Robotics}, vol.~33, no.~5, pp. 1248--1254, 2017.

\bibitem{Schouwenaars2006}
T.~Schouwenaars, E.~Feron, and J.~How, ``Multi-vehicle path planning for
  non-line of sight communication,'' in \emph{American Control Conference
  (ACC)}, 2006, pp. 5757--5762.

\bibitem{Derbas2014}
A.~M. Derbas, K.~M. Al-Aubidy, M.~M. Ali, and A.~W. Al-Mutairi, ``Multi-robot
  system for real-time sensing and monitoring,'' in \emph{15th International
  Workshop on Research and Education in Mechatronics (REM)}, 2014, pp. 1--6.

\bibitem{Varadharajan2020}
V.~S. Varadharajan, D.~St-Onge, B.~Adams, and G.~Beltrame, ``Swarm relays:
  Distributed self-healing ground-and-air connectivity chains,'' \emph{IEEE
  Robotics and Automation Letters}, vol.~5, no.~4, pp. 5347--5354, 2020.

\bibitem{Majcherczyk2018}
N.~Majcherczyk, A.~Jayabalan, G.~Beltrame, and C.~Pinciroli, ``Decentralized
  connectivity-preserving deployment of large-scale robot swarms,'' in
  \emph{IEEE/RSJ International Conference on Intelligent Robots and Systems
  (IROS)}, 2018, pp. 4295--4302.

\bibitem{Luo2020}
W.~Luo, S.~Yi, and K.~Sycara, ``Behavior mixing with minimum global and
  subgroup connectivity maintenance for large-scale multi-robot systems,'' in
  \emph{IEEE International Conference on Robotics and Automation (ICRA)}, 2020,
  pp. 9845--9851.

\bibitem{Yi2021}
S.~Yi, W.~Luo, and K.~Sycara, ``Distributed topology correction for flexible
  connectivity maintenance in multi-robot systems,'' in \emph{IEEE
  International Conference on Robotics and Automation (ICRA)}, 2021, pp.
  8874--8880.

\bibitem{Goldsmith2005}
A.~Goldsmith, \emph{Wireless Communications}.\hskip 1em plus 0.5em minus
  0.4em\relax Cambridge University Press, 2005.

\bibitem{Esposito2006}
J.~Esposito and T.~Dunbar, ``Maintaining wireless connectivity constraints for
  swarms in the presence of obstacles,'' in \emph{IEEE International Conference
  on Robotics and Automation (ICRA)}, 2006, pp. 946--951.

\bibitem{Anisi2010}
D.~A. Anisi, P.~Ögren, and X.~Hu, ``Cooperative minimum time surveillance with
  multiple ground vehicles,'' \emph{IEEE Transactions on Automatic Control},
  vol.~55, no.~12, pp. 2679--2691, 2010.

\bibitem{Boldrer2021}
M.~Boldrer, P.~Bevilacqua, L.~Palopoli, and D.~Fontanelli, ``Graph connectivity
  control of a mobile robot network with mixed dynamic multi-tasks,''
  \emph{IEEE Robotics and Automation Letters}, vol.~6, no.~2, pp. 1934--1941,
  2021.

\bibitem{Feng2015}
Z.~Feng, C.~Sun, and G.~Hu, ``Robust connectivity preserving rendezvous of
  multi-robot systems under unknown dynamics and disturbances,'' in \emph{IEEE
  Conference on Decision and Control (CDC)}, 2015, pp. 4266--4271.

\bibitem{Paolo2013}
P.~R. Giordano, A.~Franchi, C.~Secchi, and H.~H. Bülthoff, ``A passivity-based
  decentralized strategy for generalized connectivity maintenance,'' \emph{The
  International Journal of Robotics Research}, vol.~32, no.~3, pp. 299--323,
  2013.

\bibitem{zavlanos2008distributed}
M.~M. Zavlanos and G.~J. Pappas, ``Distributed connectivity control of mobile
  networks,'' \emph{IEEE Transactions on Robotics}, vol.~24, no.~6, pp.
  1416--1428, 2008.

\bibitem{zavlanos2011graph}
M.~M. Zavlanos, M.~B. Egerstedt, and G.~J. Pappas, ``Graph-theoretic
  connectivity control of mobile robot networks,'' \emph{Proceedings of the
  IEEE}, vol.~99, no.~9, pp. 1525--1540, 2011.

\bibitem{Wang2016}
L.~Wang, A.~D. Ames, and M.~Egerstedt, ``Multi-objective compositions for
  collision-free connectivity maintenance in teams of mobile robots,'' in
  \emph{IEEE 55th Conference on Decision and Control (CDC)}, 2016, pp.
  2659--2664.

\bibitem{Caregnato2022}
A.~Caregnato-Neto, M.~R. O.~A. Maximo, and R.~J.~M. Afonso, ``Resilient robust
  connectivity for multiagent systems with line of sight using mixed-integer
  programming,'' \emph{Journal of Control, Automation and Electrical Systems},
  vol.~33, no.~1, pp. 129--140, Feb 2022.

\bibitem{Gammell2014}
J.~D. Gammell, S.~S. Srinivasa, and T.~D. Barfoot, ``Informed rrt*: Optimal
  sampling-based path planning focused via direct sampling of an admissible
  ellipsoidal heuristic,'' in \emph{IEEE/RSJ International Conference on
  Intelligent Robots and Systems}, 2014, pp. 2997--3004.

\bibitem{Chen2022-1}
Y.~Chen, M.~Guo, and Z.~Li, ``Deadlock resolution and recursive feasibility in
  mpc-based multi-robot trajectory generation,'' \emph{arXiv preprint
  arXiv:2202.06071}, 2022.

\bibitem{Chen2022-2}
Y.~Chen, C.~Wang, M.~Guo, and Z.~Li, ``Multi-robot trajectory planning with
  feasibility guarantee and deadlock resolution: An obstacle-dense
  environment,'' \emph{IEEE Robotics and Automation Letters}, vol.~8, no.~4,
  pp. 2197--2204, 2023.

\bibitem{Burden2015}
R.~L. Burden, J.~D. Faires, and A.~M. Burden, \emph{Numerical analysis}.\hskip
  1em plus 0.5em minus 0.4em\relax Cengage learning, 2015.

\bibitem{Boyd2004}
S.~Boyd, S.~P. Boyd, and L.~Vandenberghe, \emph{Convex optimization}.\hskip 1em
  plus 0.5em minus 0.4em\relax Cambridge university press, 2004.

\bibitem{cvxopt}
A.~Martin, D.~Joachim, and V.~Lieven, ``Cvxopt,'' Website,
  \url{http://cvxopt.org/}.

\bibitem{Goodrich2015}
M.~T. Goodrich and R.~Tamassia, \emph{Algorithm design and applications}.\hskip
  1em plus 0.5em minus 0.4em\relax Wiley Hoboken, 2015, vol. 363.

\bibitem{cvxpy}
S.~Diamond and S.~Boyd, ``{CVXPY}: {A} {P}ython-embedded modeling language for
  convex optimization,'' \emph{Journal of Machine Learning Research}, vol.~17,
  no.~83, pp. 1--5, 2016.

\end{thebibliography}

\end{document}